\documentclass[11pt]{article}

\usepackage{amsmath,amssymb,amsthm}
\usepackage{mathtools}

\usepackage[noend]{algorithmic}
\usepackage[ruled,vlined]{algorithm2e}
\usepackage{placeins}
\usepackage{setspace}
\usepackage{url}
\usepackage{makeidx}
\usepackage{enumerate}

\usepackage{mathrsfs}
\DeclareSymbolFont{rsfs}{U}{rsfs}{m}{n}
\DeclareSymbolFontAlphabet{\mathscrsfs}{rsfs}

\def\R{{\mathbb R}}
\def\E{{\mathbb E}}
\def\P{{\mathbb P}}

\def\hE{\hat{\mathbb E}}
\def\hR{\hat{R}}

\def\naturals{{\mathbb N}}

\def\bfone{{\boldsymbol 1}}
\def\da{{\partial a}}
\def\Ball{{\sf B}}

\def\hm{\hat{m}}

\def\btheta{{\boldsymbol \theta}}
\def\id{{\boldsymbol I}}
\def\bg{{\boldsymbol g}}
\def\bm{{\boldsymbol m}}
\def\bx{{\boldsymbol x}}
\def\by{{\boldsymbol y}}

\def\bX{{\boldsymbol X}}
\def\bY{{\boldsymbol Y}}
\def\bW{{\boldsymbol W}}
\def\bA{{\boldsymbol A}}

\def\bzero{{\boldsymbol 0}}

\def\bT{{\boldsymbol T}}
\def\bB{{\boldsymbol B}}

\def\ox{{\overline{x}}}
\def\obx{{\overline{\boldsymbol x}}}

\def\ord{{\sigma}}

\def\bzero{{\boldsymbol 0}}

\def\normal{{\sf N}}

\def\<{{\langle}}
\def\>{{\rangle}}

\def\ssat{\mbox{\tiny\rm s}}
\def\sd{\mbox{\tiny\rm d}}
\def\smllc{\mbox{\tiny\rm c}}
\def\sSAT{\mbox{\tiny\rm SAT}}
\def\sXOR{\mbox{\tiny\rm XOR}}

\def\KL{{\sf KL}}
\def\mask{\mbox{\tiny\rm mask}}
\def\diff{\mbox{\tiny\rm diff}}
\def\sfree{\mbox{\tiny\rm free}}

\def\BPc{{\rm BP}_{\mbox{\rm\footnotesize c}}}
\def\BPu{{\rm BP}_{\mbox{\rm\footnotesize u}}}

\def\BP{\mathrm{BP}}

\def\ns{n_{\mbox{\rm\footnotesize s}}}

\def\Unif{{\sf Unif}}
\def\cS{{\mathcal S}}
\def\cL{{\mathcal L}}

\def\eps{{\varepsilon}}

\def\sP{{\sf{P}}}

\def\hQ{\widehat{Q}}
\def\bfone{{\boldsymbol 1}}

\newcommand{\sign}{\mathrm{sign}}

\newcommand{\beq}{\begin{equation}}
\newcommand{\eeq}{\end{equation}}

\renewcommand{\P}{\mathbb{P}}

\newcommand{\bu}{\mathrm{\bf u}}
\newcommand{\bv}{\mathrm{\bf v}}

\newcommand{\bb}{\text{\boldmath $b$}}

\newcommand{\cM}{\mathcal{M}}

\def\normal{{\sf N}}
\def\id{{\boldsymbol I}}

\def\de{{\rm d}}

\def\sTV{\mbox{\scriptsize\rm TV}}

\def\eps{\varepsilon}

\newcommand{\hmu}{\widehat\mu}

\newcommand{\atanh}{\operatorname{atanh}}

\def\bx{{\boldsymbol x}}

\def\bfone{{\boldsymbol 1}}

\def\cQ{{\cal Q}}
\def\hcQ{\widehat{\cal Q}}
\def\Poisson{{\sf Poisson}}

\usepackage[top=1in, bottom=1.25in, left=1in, right=1in]{geometry}
\usepackage[utf8]{inputenc}

\usepackage{graphicx}
\usepackage{float}
\usepackage{psfrag}
\usepackage{epsfig}
\usepackage{caption}
\usepackage{subfig}
\usepackage[usenames,dvipsnames,svgnames,table]{xcolor}
\usepackage{epstopdf}
\usepackage{color}

\usepackage[
  pagebackref,
  colorlinks=true,
  pdfpagemode=UseNone,
  citecolor=OliveGreen,
  linkcolor=BrickRed,
  urlcolor=BrickRed,
  pdfstartview=FitH
]{hyperref}

\usepackage{xr}
\usepackage{cancel}
\usepackage{calc}
\usepackage{scalerel,stackengine}
\usepackage[numbers,sort&compress]{natbib}
\stackMath
\newcommand\reallywidehat[1]{%
\savestack{\tmpbox}{\stretchto{%
  \scaleto{%
    \scalerel*[\widthof{\ensuremath{#1}}]{\kern.1pt\mathchar"0362\kern.1pt}%
    {\rule{0ex}{\textheight}}%
  }{\textheight}%
}{2.4ex}}%
\stackon[-6.9pt]{#1}{\tmpbox}%
}

\numberwithin{equation}{section}

\newtheoremstyle{myexample}
    {\topsep}
    {\topsep}
    {\rm}
    {}
    {\bf}
    {.}
    {.5em}
    {}

\newtheoremstyle{myremark}
    {\topsep}
    {\topsep}
    {\rm}
    {}
    {\bf}
    {.}
    {.5em}
    {}

\newtheorem{claim}{Claim}[section]

\newtheorem{theorem}{Theorem}
\newtheorem{proposition}[claim]{Proposition}

\theoremstyle{myremark}

\theoremstyle{myexample}

\def\ti{t}
\def\tdi{t}

\definecolor{darkgreen}{rgb}{0.0, 0.5, 0.0}

\title{Generating from Discrete Distributions Using Diffusions:\\
Insights from Random Constraint Satisfaction Problems}

\author{
Alankrita Bhatt$^{1}$\thanks{Authors are listed in alphabetical order.}
\and
Mukur Gupta$^{1}$\footnotemark[1]
\and
Germain Kolossov$^{1}$\footnotemark[1]
\and
Andrea Montanari$^{1,2}$\footnotemark[1]
}

\date{
$^{1}$Granica Computing Inc. -- \href{https://www.granica.ai/}{granica.ai}\\
$^{2}$Stanford University
}

\begin{document}

\maketitle

\begin{abstract}
Generating data from discrete distributions is important
for a number of application domains including text, 
tabular data, and genomic data. Several groups have recently
used random $k$-satisfiability ($k$-SAT) as a synthetic 
benchmark for new generative techniques. In this paper, we show that fundamental insights from the 
theory of random constraint satisfaction problems have observable implications
(sometime contradicting intuition) on the behavior of generative techniques on 
such benchmarks. 
More precisely, we study the problem of generating a uniformly random solution
of a given (random) $k$-SAT or $k$-XORSAT formula. Among other findings, we observe that:
$(i)$~Continuous diffusions outperform masked discrete diffusions; $(ii)$~Learned diffusions 
can match the theoretical `ideal' accuracy; $(iii)$~Smart ordering of the variables can significantly
improve accuracy, although not following popular heuristics.
\end{abstract}

\FloatBarrier
\section{Introduction} \label{sec:intro}

Since the introduction of denoising diffusions 
\cite{sohl2015deep,song2019generative,ho2020denoising,song2021score},
an impressive number of variants have emerged, in particular to generate
data over discrete domains, most notably language and tabular data 
\cite{austin2021structured,hoogeboom2021argmax,shi2024simplified}.
Despite this, the relative merits of different techniques 
for generating discrete data remain poorly understood. This is particularly true for tabular data generation
\cite{liu2023goggle,zhang2024mixed,shi2024tabdiff},
an application domain that lacks a universally representative set of 
benchmarks.

Several groups have recently used random $k$-satisfiability ($k$-SAT) as a synthetic benchmark for discrete data generation \cite{selsam2019learning,freivalds2022denoising,ye2025beyond,anil2025interleaved}. 
However, the choice of the setting and 
metrics in these works is somewhat arbitrary, and not guided by fundamental principles.
In particular, there is no notion of what is the target accuracy the generative model should strive for,
and no guidance about the parameters of the $k$-SAT distribution to be used.

The objective of this paper is twofold. \emph{First,} 
we will show that the theory of random constraint satisfaction
problems (CSPs) can provide guidance about the specific setting 
(specific CSP and parameters) as well as target accuracy for benchmarking generative
techniques. \emph{Second,} we will show that insights from that theory 
are predictive of the behavior of diffusion sampling techniques, 
leading to conclusions that can contradict naive intuition.

\begin{figure}[t]
    \centering
    \includegraphics[width=0.93\linewidth]{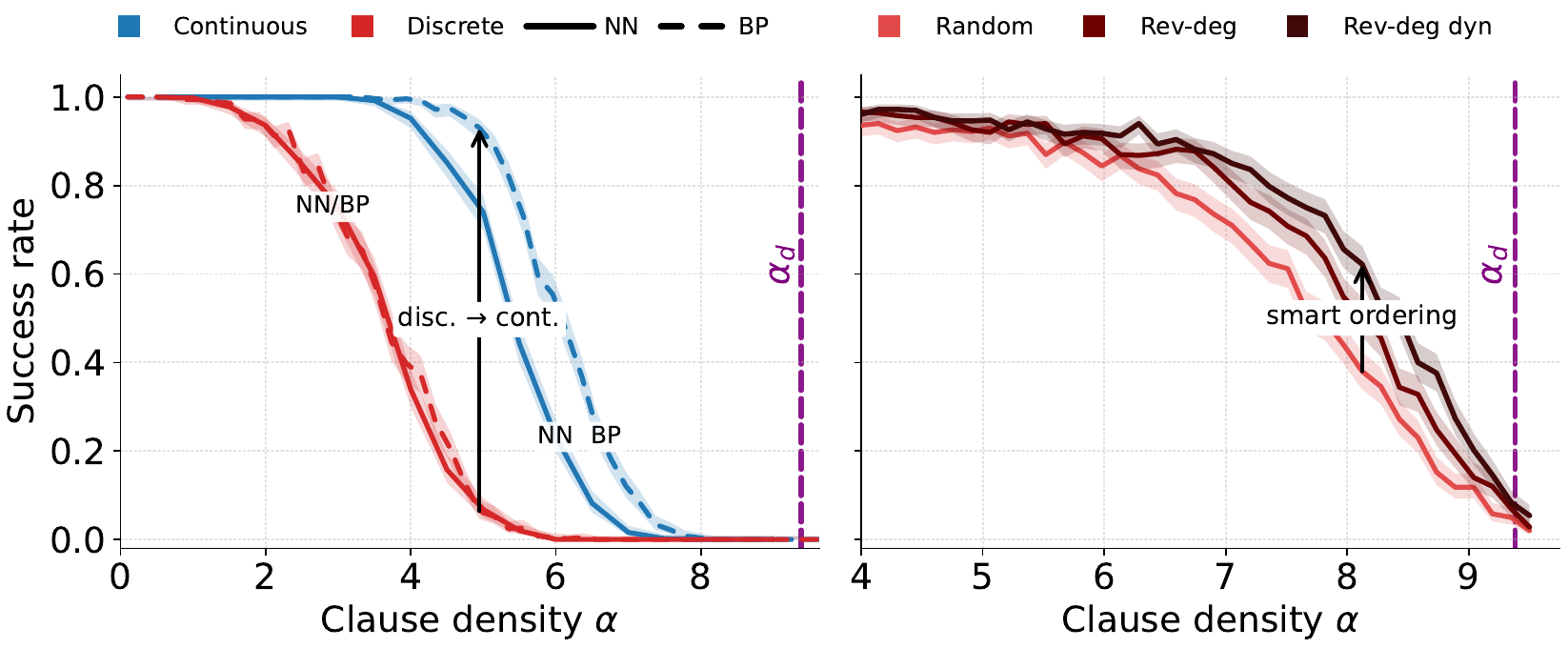}
    
    \caption{
        \textbf{4-SAT ($N=300$)} Left: The learned NN denoiser (solid) matches the BP denoiser (dashed), and continuous diffusion (blue) outperforms discrete diffusion (red) for both denoisers at locality radius $r=3$. 
        Right: Smart variable ordering improves discrete diffusion with $r=100$ BP denoiser.
        We compare random ordering with reverse-degree and dynamic reverse-degree orderings. We report success rate (probability of producing a satisfying assignment) versus clause density $\alpha$, averaged over 500 random formulas. 
        Shaded regions: 95\% confidence interval.
        Vertical line: dynamical phase transition $\alpha_{\sd}(k=4)\approx 9.38$.}\vspace{-0.3cm}
    \label{fig:fig1_summary_ksat}
\end{figure}

The present work builds upon the wealth of mathematical 
understanding of random CSPs that has been developed over the last twenty years,
through an interdisciplinary effort across computer science, statistical physics, and probability theory,
see \cite{mezard2002analytic,krzakala2007gibbs,ding2022proof,ayre2020satisfiability,MezardMontanari} 
for a few pointers.
Recently, methods from this literature were used for construct algorithms that provably 
sample from high-dimensional and highly multi-modal Gibbs measures 
arising from spin glass models  or random CSPs
\cite{el2022sampling,montanari2023posterior,huang2024sampling,ghio2024sampling,mei2025deep} 
(see also \cite{montanari2007solving}
for an earlier example).
These algorithms implement denoising diffusions  or discrete diffusions,
with the difference that the denoiser 
is constructed explicitly by using (variants of) loopy belief propagation (BP).
For large random instances, loopy BP  is expected to provide a near-optimal 
denoiser among polynomial-time algorithms.

Here, we take this line of work one step forward,
by investigating learned diffusions and comparing them with the behavior predicted from 
the theory for BP-based diffusions.

An instance of a binary CSP is given by a set of constraints over the 
vector $\bx\in\{+1,-1\}^N$.
A solution of a specific CSP instance is a vector $\bx$ that satisfies all the constraints.
We attempt to generate samples from the uniform measure over solutions of a given CSP,
using both standard denoising diffusions and masked discrete diffusions.
For each of these two approaches and each of four families of
random CSPs ($k$-SAT, $k$-XORSAT, $k$-NAESAT and hypergraph bicoloring), 
we study the behavior of generative methods based on several learned and
BP-based denoisers. (We use the term `denoiser' broadly:
for masked discrete diffusions this is the conditional probability model.)

We succinctly summarize some of our findings below, and refer to Figure \ref{fig:fig1_summary_ksat} for an illustration:

\emph{1. Continuous diffusions outperform masked discrete diffusions.} 
One of the main motivation for the introduction of discrete diffusions 
was the idea that they are better adapted to discrete data \cite{hoogeboom2021argmax,austin2021structured}. 
We show that this is not
the case for random CSPs, in agreement with  asymptotic theory
\cite{montanari2007solving,ghio2024sampling}.

\emph{2. Ideal target accuracy from random CSP theory.} We mostly focus on 
`local' denoisers, i.e. denoisers that estimate each variable using a bounded radius 
neighborhood of that variable in the graph $G$ which represent the CSP instance.  
We derive a prediction for the ideal `local' denoiser, i.e. the denoiser that minimizes
the score matching objective in the large size, large sample size limit.

\emph{3. Learned diffusion match theoretical target.}
We show empirically that learned local denoisers can match the behavior of 
the ideal denoisers at the previous point.

\emph{4. Accuracy is improved by smart ordering (continuous and discrete).}
We study  a more general class of generative methods which treats different 
coordinates $x_i$ in a different way. In the case of
discrete diffusions, this corresponds to unmasking variables in a smart order.
For continuous diffusions, this corresponds to adding noise with different variances
to different variables. We construct schemes that significantly improve
the success rate of both discrete and continuous diffusions.
\begin{figure}[t]
    \centering
        \includegraphics[width=0.97\linewidth]{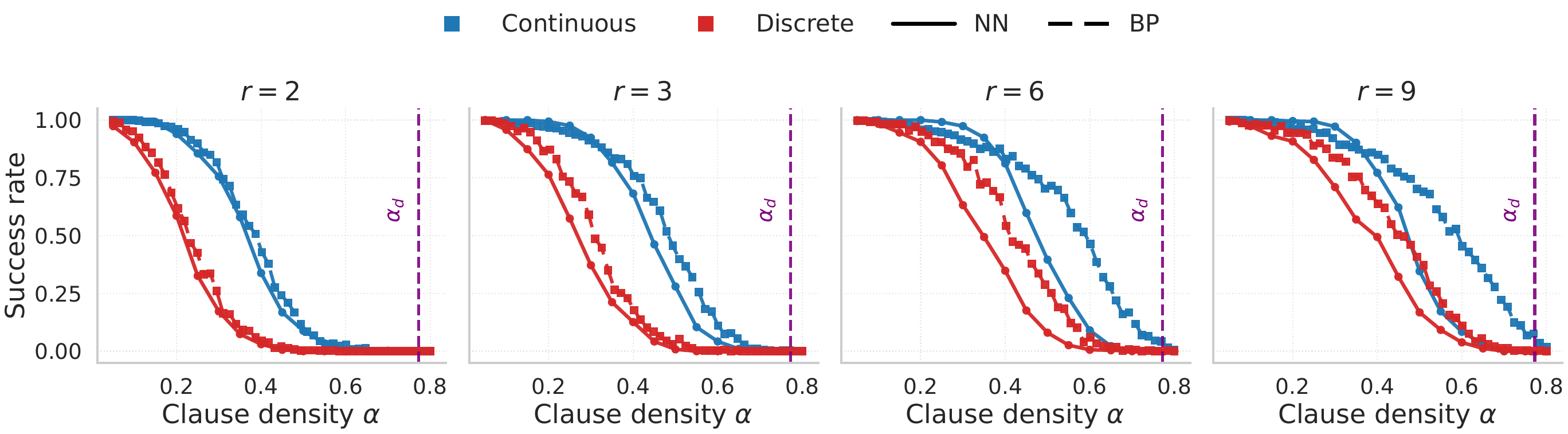}
    \caption{
        \textbf{4-XORSAT ($N=100$): learned NN denoiser (solid lines) vs. BP denoiser (dashed lines).}
        Success rate as a function of clause density~$\alpha$ for continuous (blue) and discrete (red) diffusions.
        Each panel corresponds to a different locality radius for the denoisers~$r \in \{2,3,6,9\}$.
        Success rates are computed over 500 random formulas. The vertical line corresponds
        to the dynamical phase transitoion at 
        $\alpha_{\sd}(k=4)\approx 0.77228$. 
    }
    \label{fig:kxorsat_diffusion_vs_bp}
\end{figure}

\section{Setting and background}
\label{sec:background}

We consider four families of random CSPs with 
binary decision 
%
$\bx=(x_1,\dots,x_N)\in\{+1,-1\}^N$:
$k$-SAT, $k$-XORSAT, $k$-NAE-SAT, and $k$-hypergraph bicoloring.  An instance is defined by a set 
$M$ constraints over the $N$ variables. Each constraint (`clause') comprises $k$
variables (or their negation).

\noindent{\bf $k$-SAT.} Constraint $a$ takes the form
$(z_{i_a(1)}\vee z_{i_a(2)}\vee \cdots\vee z_{i_a(k)})=+1$ where 
$i_a(1),\dots,i_a(k)\subseteq [N]$ are distinct, $z_{i_a(j)} = s_{a,j} x_{i_a(j)}$ with
$s_{a,j}\in\{+1,-1\}$ and $x \vee x'=\max(x,x')$  
(logical OR in the True$/$False encoding).

\noindent{\bf $k$-XORSAT.} Constraint $a$ takes the form 
$x_{i_a(1)}\cdot x_{i_a(2)}\cdots x_{i_a(k)}=s_a$, where $s_{a}\in\{+1,-1\}$,
and $\cdot$ denotes simple product.

Because of space limitations, we defer definitions and results for 
{\bf NAE-SAT}, and {\bf hypergraph bicoloring} to the appendices. (Our results
for SAT and XORSAT are more complete, and confirmed by the additional examples in the appendices.)
A random $k$-SAT/XORSAT formula is generated by drawing each of the $M$ 
constraints independently and uniformly at random among the $2^k\binom{N}{k}$ 
possible ones (for $k$-SAT) or $2\binom{N}{k}$
possible ones (for $k$-XORSAT).   We will denote by $\da:=\{i_a(1),\dots, i_a(k)\}$
the set of variable indices appearing in constraint $a$.

$k$-SAT and $k$-XORSAT belong to different
complexity classes.
Because of the linear-algebraic 
structure,
a XORSAT solution can be found in
polynomial time, and indeed it is easy to sample a uniformly random solution. 
In contrast, even simply finding a $k$-SAT is NP complete 
\cite{arora2009computational}. We emphasize that XORSAT is a challenging testing ground
nevertheless because: $(i)$~We limit ourselves to local algorithms for which SAT and 
XORSAT are much more similar; $(ii)$ Denoising diffusions require solving a denoising problem that
is worst-case hard also for XORSAT \citep{ghazi2017lp} .

An instance of $k$-SAT or $k$-XORSAT can be encoded in a factor graph 
$G=(V,F,E)$, with variable nodes $V=[N]$ (corresponding to the $N$ variables), 
factor nodes $F$, $|F|=M$ (corresponding to the $M$ constraint), and edges $(i,a)$, $i\in V$, $a\in F$,
every time variable $x_i$ appears in constraint $a$. $G$ is a marked graph, with marks on the edges
carrying information of
the signs $s_{a,j}$, $s_a$. Given
an instance $G$, we are interested in the uniform measure over solutions
\begin{align}
\mu_G(\bx) = \frac{1}{Z_G}\bfone_{\{\bx \mbox{ solves } G\}} = \frac{1}{Z_G} \prod_{a\in F}\psi_a(\bx_{\da})\, ,
\end{align}
where $Z_G$ is the number of solution of formula $G$, and $\psi_a(\bx_{\da})=1$
 if $\bx$
satisfies the $a$-th constraint, while $\psi_a(\bx_{\da})=0$ otherwise. We consider two sampling schemes.

\noindent{\bf Denoising diffusions.} We update a vector $\bY_\ell\in\R^N$,
$\ell\in\{0,1,\dots,L\}$, with initialization $\bY_0\sim\normal(0,\id_N)$, and proceeding
according to the Euler discretization \cite{ho2020denoising,song2019generative}
\begin{align}
\bY_{\ell+1}= \gamma_{\ell}\bY_\ell +\delta_\ell \bW\, \bm\big(\bY_\ell;\ti_{\ell}\big)+\sqrt{\beta_\ell}\bg_\ell\, ,
\label{eq:FirstDiff}
\end{align}
where $(\bg_\ell)_{\ell\ge 0}\sim_{iid}\normal(\bzero,\id_N)$, $\beta_\ell\in (0,1)$ are stepsizes, 
$\ti_{\ell}=\prod_{i=\ell}^{L-1}(1-\beta_i)$, $\gamma_\ell = (1-\beta_\ell/(1-\ti_\ell))/\sqrt{1-\beta_\ell}$,
$\delta_{\ell} = \beta_{\ell}\sqrt{\ti_{\ell}}/((1-\ti_\ell)\sqrt{1-\beta_\ell})$. 
(Here $\ell$ increases along the generation process.)

Further, $\bW$ is a matrix which we will take diagonal
and non-negative, and 
can be interpreted as a weighting scheme determining the order in which information about
variables is revealed.
Finally, $\bm(\,\cdot\,)$ is the denoiser. The ideal denoiser 
$\bm(\by;\ti)$ is
the posterior mean given observations 
$\by = \sqrt{\ti}\bW\bx+ \sqrt{1-\ti}\, \bg$, $\bg\sim\normal(0,\id_N)$. 
This coincides with the mean of the tilted measure 
\begin{align}
\mu_G^{\by,\ti}(\bx) = \frac{1}{Z_G(\ti)} \prod_{a\in F}\psi_a(\bx_{\da}) \,e^{\sqrt{\ti}\<\by,\bW\bx\>/(1-\ti)}\, ,\label{eq:Tilted}
\end{align}
for a suitable normalizing constant $Z_G(\ti)$.
%
A key observation is that the tilted measure \eqref{eq:Tilted} factors according to
the same graph $G=(V,F,E)$ as the original measure $\mu_G$. In particular, we can employ 
BP to attempt to approximate the marginals of $\mu_G^{\by,\ti}$ and hence the denoiser $\bm(\,\cdot\,)$.

\noindent{\bf Discrete masked diffusions.} It is useful to 
introduce the notation $\bx_{a:b}=(x_a,x_{a+1},\dots, x_b)$ for $a\le b$ (and 
$\bx_{:b}=\bx_{1:b}$, $\bx_{a:}=\bx_{a:N}$). Given an arbitrary permutation
$\ord:[N]\to [N]$ of the variables, we let $\bx^{\ord}=(x_{\ord(1)},\dots , x_{\ord(N)})$
be the variables ordered according to $\ord$, and 
\begin{align}\label{eq:ConditionalDistr}
\mu_G^{\ord,\ell}(\bx|\obx^\ord_{1:\ell})
=\P_{\bX\sim\mu_G}\big(\bX=\bx\big|\bX^{\ord}_{1:\ell}=\obx^\ord_{1:\ell}\big)\, ,
\end{align}
the conditional distribution given assignments of the first $\ell$ variables  
to $\obx^\ord_{1:\ell}$. With a slight abuse of notation, we also
write $\mu_G^{\ord,\ell}(x^{\ord}_{\ell+1}|\bx^\ord_{1:\ell})$ 
for the conditional distribution of $x^{\ord}_{\ell+1}$ given
$\obx^\ord_{1:\ell}$. We then sequentially sample variables by repeating for
$\ell\in\{0,\dots, N-1\}$, i.e. $\ox_{\ord(\ell+1)}\sim \mu_G^{\ord,\ell}(\,\cdot\, |\obx^\ord_{1:\ell})\,$.
In the following, the order of variables $\ord$ will be chosen 
to be purely as a function of the instance
$G$.

We note that the conditional distribution \eqref{eq:ConditionalDistr}
also factors according to the graph $G$, namely
\begin{align}\label{eq:CondMarkov}
\mu_G^{\ord,\ell}(\bx|\obx^\ord_{1:\ell})= \frac{1}{Z_{G,\ell}}
\prod_{a\in F}\psi_a(\bx_{\da}) \prod_{i=1}^{\ell}\bfone_{x_{\ord(i)}=\ox_{\ord(i)}}\, .
\end{align}

\noindent{\bf Local denoisers.} Implementing either continuous or discrete diffusions
requires to compute marginals of probability distributions that factor according to the graph 
$G$, respectively, the distribution $\nu(\bx)=\mu_G^{\by,\ti}(\bx)$ of Eq.~\eqref{eq:Tilted}
or $\nu(\bx)=\mu_G^{\ord,\ell}(\,\cdot\, |\obx^\ord_{1:\ell})$ of Eq.~\eqref{eq:CondMarkov}.
We refer to any any function that take as input $(G,\by,\ti)$ (in the first case) or $(G,\ord,\obx^\ord_{1:\ell})$
(in the second)
and computes the marginal $\nu(x_i\in \,\cdot\,)$ of this distribution as to a `denoiser'
(hence this term refers both to continuous and discrete diffusions).
We will focus on denoisers that are $r$-\emph{local} with respect to $G$, 
in the sense that the marginal at $i\in V$ is estimated by a function of the radius-$r$ neighborhood of $i$.

Namely, let $\Ball_G(i,r)$ be the subgraph of $G$ induced by
vertices $j\in V$ such that $d_G(i,j)\le r$ (with $d_G$ the graph distance, s.t. if $i,j\in V$
belong to the same constraint $d_G(i,j)=1$). A denoiser is $r$-local if
the marginal computed at $i\in V$ only depends on $(G,\by)$  or $(G,\obx^\ord_{1:\ell})$
via their restriction to $\Ball_G(i,r)$. We denote the information in such a neighborhood
by $\bY_{\Ball_G(i,r)}$, e.g., for continuous diffusions,
$\bY_{\Ball_G(i,r)} :=(i,\Ball_G(i,r),\by\big|_{\Ball_G(i,r)})$.
We will use two classes of local denoisers:

\noindent\emph{(i) BP denoiser.} Since the probability distributions $\mu_G^{\by,\ti}(\bx)$ and $\mu_G^{\ord,\ell}(\,\cdot\, |\obx^\ord_{1:\ell})$
factorize according to the graph $G$, we can use (loopy) BP to approximate their marginals,
see Appendix \ref{appendix:BPDetails} for details.
Provided we use an initialization that is independent 
of $G$, $r$ iterations of BP result in an $r$-local denoiser. We will consider two initialization `uniform' and `cavity,'
and denote the corresponding denoisers by $\BPu(r)$, $\BPc(r)$.
In the first one, we initialize all beliefs to be uniform  on $\{+1,-1\}$,
while the second one will be defined in Section \ref{sec:Cavity}.

\noindent\emph{(ii) Neural Network (NN) denoiser.} We construct a neural network 
architecture 
which takes as input the radius $r$ neighborhood of an arbitrary vertex $i$ 
and information revealed within such a neighborhood, 
and returns the predicted logit of variable $x_i$:
$f(\,\cdot\, ;\btheta):\; \bY_{\Ball_G(i,r)}\;\mapsto \; f(\bY_{\Ball_G(i,r)};\btheta)\; \;\in \R$.

Such a model also induces a predicted conditional expectation for variable $i$ 
given information $\bY_{\Ball_G(i,r)}$,
via $\hm(\bY ;\btheta) = \tanh(f(\bY;\btheta)/2)$. This 
corresponds to a unique predicted
probability distribution $\nu_i$ on $\{+1,-1\}$.
Our architecture has parameters $\btheta$ and is --by construction-- equivariant with respect to relabeling of
the vertices.
We refer to Appendix 
\ref{appendix:NNArchitrecture}
for a description of this architecture.

\noindent{\bf Satisfiability and dynamical phase transitions.} As $N, M\to\infty$, with constant number of constraints per variable
$M/N=\alpha$, the space of solutions of random $k$-SAT/XORSAT undergoes certain phase transitions\footnote{This picture is rigorously established for $k$-XORSAT \cite{ibrahimi2015set,ayre2020satisfiability} and, in a sequence of recent mathematical breakthroughs
\cite{ding2022proof,coja2023rigorous}, for $k$-SAT at $k$ above a certain constant. It is still conjectural for --say-- $k=3,4$, see  \cite{MezardMontanari} the conjectured picture.}.
Two phase transitions at locations $\alpha_{\sd}<\alpha_{\ssat}$ are particularly relevant (their location 
differ between SAT and XORSAT, but we omit this dependence unless useful):

\noindent$(1)$ The \emph{satisfiability phase transition} at $\alpha_{\ssat}(k)$: for $\alpha<\alpha_{\ssat}(k)$
a random instance $G$ has solutions with high probability (whp, i.e. with probability converging to one as $N,M\to\infty$);
for  $\alpha>\alpha_{\ssat}(k)$ it does not,  whp. (For $\alpha<\alpha_{\ssat}(k)$, the number of solutions is exponentially large in $N$.)

\noindent$(2)$ The \emph{dynamical (or, shattering) phase transition} at $\alpha_{\sd}(k)$: for $\alpha<\alpha_{\sd}(k)$
the set of solutions $\cS\subseteq\{\pm 1\}^N$ is whp `well connected' while 
for  $\alpha>\alpha_{\sd}(k)$ it comprises an exponential number of  components $\cS=\cup_{j=1}^L\cS_{j}$.
Any two distinct components have Hamming distance $d(\cS_{i},\cS_j)\ge cN$. 

The dynamical phase transition is particularly important  because it is expected
that no polynomial time algorithm can sample $k$-SAT solutions for $\alpha \in (\alpha_{\sd}(k), \alpha_{\ssat}(k))$
\cite{krzakala2007gibbs,bresler2022algorithmic}. For $k$-XORSAT, sampling is always tractable exploiting the linear-algebraic structure, but 
local algorithms are believed to fail for  $\alpha \in (\alpha_{\sd}(k), \alpha_{\ssat}(k))$.
Our objective will be to sample solutions for $\alpha\in (0,\alpha_{\sd}(k))$.

\noindent{\bf Phase transitions for diffusions.} 
For a diffusion-based sampler to be successful, it needs to have access to an oracle that compute
marginals with respect to tilted measure $\mu_G^{\by,\ti}(\,\cdot\,)$ of Eq.~\eqref{eq:Tilted}
(for continuous diffusions) or $\mu_{G}^{\ord,\ell}(\,\cdot\,)$ of Eq.~\eqref{eq:CondMarkov}
(for discrete diffusions). As a consequence, if marginal computation (or --equivalently-- sampling)
is intractable for these measures, then diffusion sampling fails, see \cite{montanari2025computational} for a formal statement.

%
The  measure $\mu_{G}^{\ord,\ell}(\,\cdot\,)$ is just the uniform measures
over the set of solutions of a modified random CSP obtained by constraining some of the variabled to be equal to the `observed' values. This modified CSP undergoes a dynamical phase transition $\alpha_{\sd}(k;t)$. It was conjectured in \cite{montanari2007solving} that discrete diffusions will succeed only if
$\alpha\le \alpha_{\mask}(k):=\inf_{t\in [0,1]}\alpha_{\sd}(k;t)$.
This threshold was computed in \cite{montanari2007solving} for SAT, 
while for XORSAT the computation of $\alpha_{\mask}(k)$ can be carried out
similarly: we have  $\alpha^{\sXOR}_{\mask}(4)\approx 0.5625$, $\alpha^{\sSAT}_{\mask}(4)\approx 8.05$. A similar threshold can be computed for continuous diffusions with $\bW=\id_N$, yielding in particular 
$\alpha^{\sXOR}_{\diff}(4)\approx 0.63$ (Appendix \ref{appendix:BPDetails}).
While a complete proof of this conjecture is still lacking, the positive direction was proven
in a related spin-glass model in \cite{huang2024sampling}. 
\section{Main results} \label{sec:mainresults}


We will compare generative methods both in theory and practice across three different 
axes: learned vs BP denoisers; discrete vs continuous diffusions; uniform vs smart variable ordering.
We will mostly report the probability that they generate  solutions (`success rate'). 
This is a relatively strict criterion since a single incorrect variable
can result in a failure (part of previous work considered the fraction of satisfied constraints, a weaker metric.)
Since high success rate does not guarantee the generated solutions to be 
uniformly distributed, we will test approximate uniformity in Section \ref{sec:Uniformity}.

We emphasize that BP denoisers do not require training. For NN denoisers, 
we train on solutions/noisy observation pairs
$\bx^{(s)},\by^{(s,\ell)}$. In the case of XORSAT  it is easy to 
generate uniform solutions using linear-algebraic methods.
%
%
%
For other CSPs, we experimented with several heuristics to generate solutions, 
as discussed in Appendix \ref{appendix:NNExptDetails}.
As already mentioned, we defer a presentation of results for NAE-SAT and hypergraph bicoloring to 
Appendix \ref{app:further_results}.

We train $r$-local NN models $f(\,\cdot;\btheta)$, by minimizing 
$\hR_n(\btheta)= \frac{1}{NL}\sum_{\ell\le L, i\le N} \hE_{n,\ell}
\cL\big(x_i,f(\bY_{\Ball_G(i,r)};\btheta)\big)\, .\label{eq:ScoreMatching}$
For continuous diffusions  we use the square loss
$\cL(x,f) = (x-\tanh(f/2))^2$ (recall that the conditional mean is related to
the log-likelihood via $\hm=\tanh(f/2)$), while for discrete masked diffusions we use 
the cross-entropy $\cL(x,f)=-xf/2+\log\cosh(f/2)$.
Here $\bY_{\Ball_G(i,r)} = (i,\Ball_G(i,r),\by\big|_{\Ball_G(i,r)})$ is the entire local
information available to the denoiser, 
$\hE_{n,\ell}$ denotes empirical expectation with respect to the training samples
$(G^{(s)}, \bx^{(s)},\by^{(s,\ell)})$, $s\in\{1,\dots, \ns\}$,  where $G^{(s)}$ is a random
$k$-SAT/XORSAT instance, $\bx^{(s)}$ is a random solution of  $G^{(s)}$, and
$\by^{(s,\ell)} = \sqrt{\ti_{\ell}}\bx^{(s)} + \sqrt{1-\ti_{\ell}}\bg^{(s,\ell)}$,
$\bg^{(s,\ell)} \sim\normal(0,\id_N)$ (for continuous diffusions) or $\obx^{\ord}_{1:\ell}$
(for discrete diffusions).

\begin{figure}[t]
    \centering
    \includegraphics[width=0.45\linewidth]{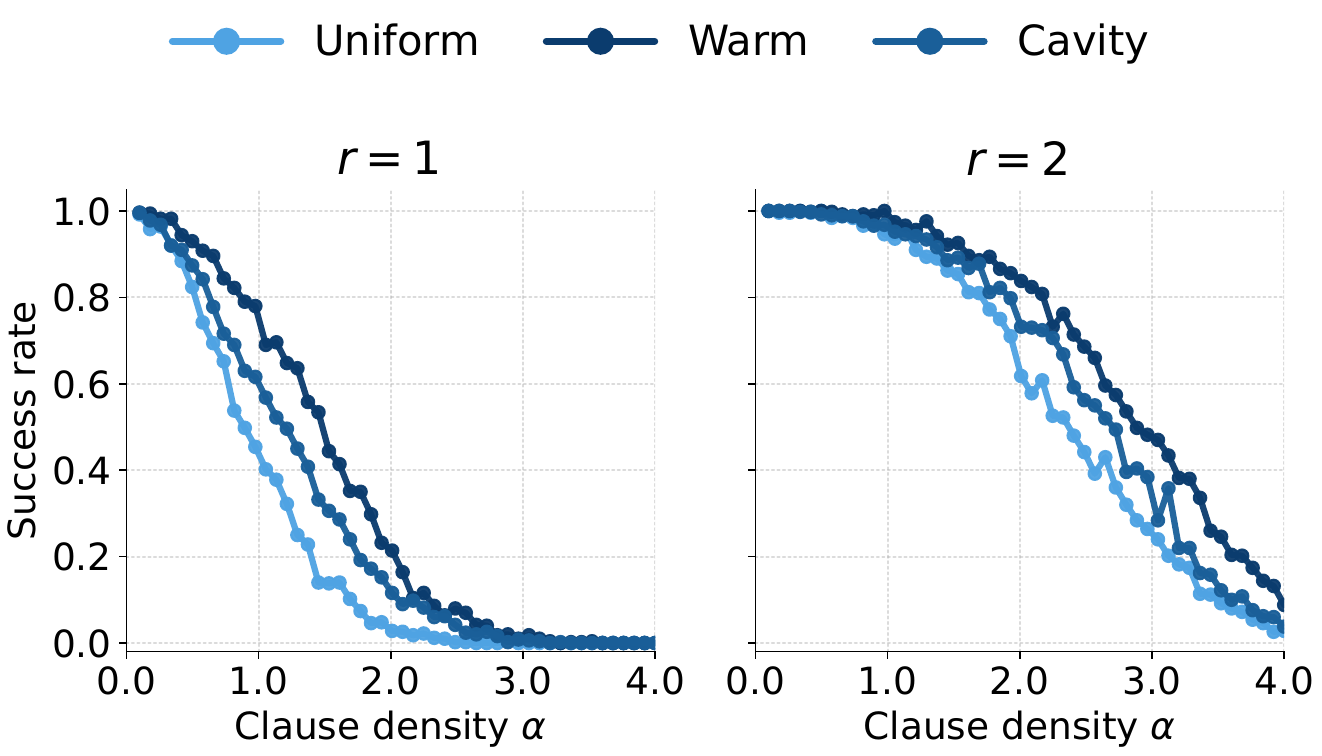}
     \phantom{AAA}
    \includegraphics[width=0.45\linewidth]{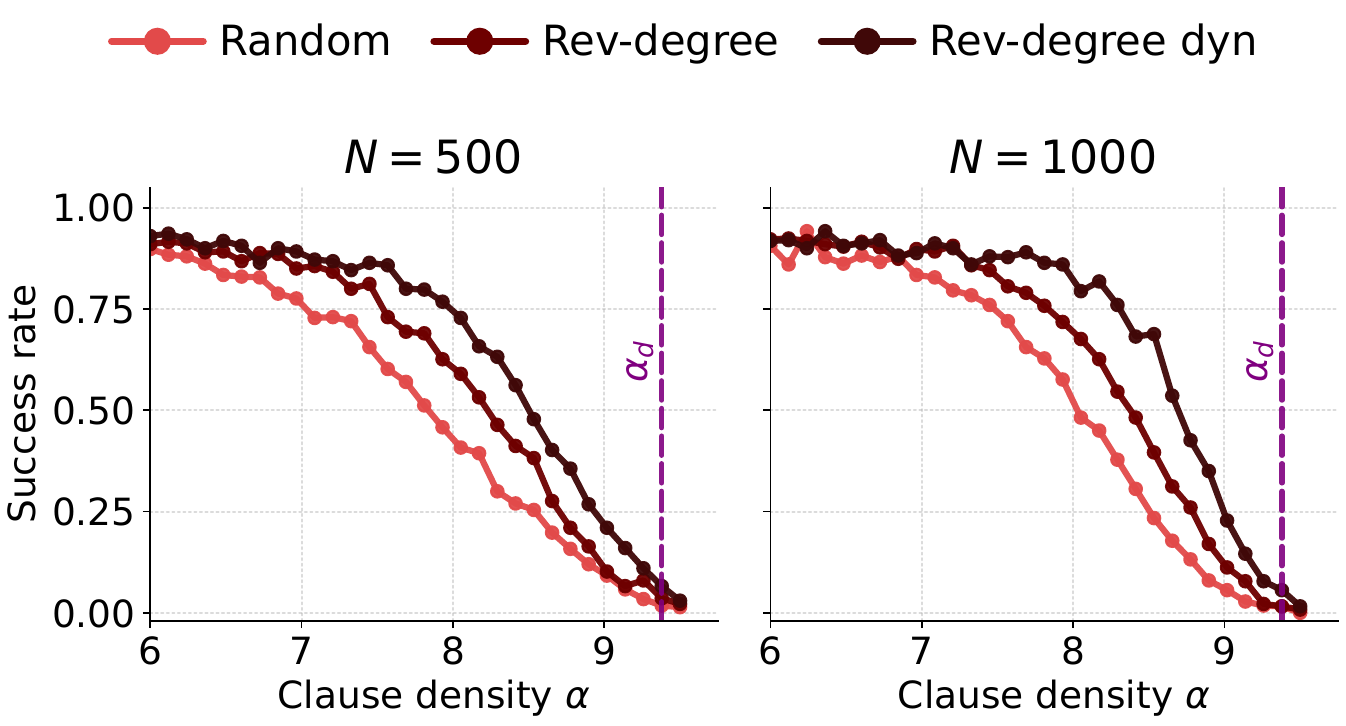}

    \caption{
        \textbf{4-SAT.} \textbf{Two left panels:} Effect of BP message initialization in discrete diffusion. We compare zero, warm-start, and cavity-based initializations for $N=300$ using a BP denoiser with radius $r=1$ (\textit{left}) and $r=2$ (\textit{right}). \textbf{Two right panels:} Effect of variable ordering in discrete diffusion with a BP denoiser. We compare random, reverse min-degree, and dynamic reverse min-degree orderings for $N=500$ (\textit{left}) and $N=1000$ (\textit{right}) with $r=100$. Each panel reports success rate versus clause density $\alpha$, averaged over 500 random formulas; the vertical line marks the dynamical phase transition $\alpha_{\sd}(k=4)\approx 9.38$.
    }
    \label{fig:bp_discrete_init_r2}
\end{figure}

\subsection{Continuous outperform 
discrete diffusions; NN matches BP} \label{sec:results_cont_vs_disc}

In Figures \ref{fig:fig1_summary_ksat} (left frame) and \ref{fig:kxorsat_diffusion_vs_bp} we compare the 
success rate of continuous and discrete diffusions, both using 
BP and NN denoisers, respectively for $4$-SAT (with $N=300$) and $4$-XORSAT (with $N=100$). 
In Figure~\ref{fig:kxorsat_diffusion_vs_bp},  each panel corresponding to a different locality radius
(see Appendix for the analogous comparison for SAT). We observe that 
continuous diffusions systematically outperforms discrete diffusions. 

The non-rigorous theoretical predictions of \cite{montanari2007solving,ghio2024sampling}
imply sharp thresholds for  discrete/continuous diffusion sampling at
$\alpha^{\sSAT}_{\mask}(4)<\alpha^{\sSAT}_{\diff}(4)$ for $4$-SAT and
$\alpha^{\sXOR}_{\mask}(4)<\alpha^{\sXOR}_{\diff}(4)$ for $4$-XOR.
This theory applies to the BP denoiser under the asymptotics $N,M\to\infty$ as well as $r\to\infty$. 
Our results demonstrate that:
$(a)$ The asymptotic theory is relevant at moderate $N,M,r$;
$(b)$ The  predictions for BP denoisers  also capture the behavior of 
learned NN denoisers; $(c)$ The success rate of learned NN denoisers nearly matches the one achieved with BP denoisers.

The last point is higly non-trivial, given that the NN architecture is only aware of the permutation invariance and locality structure
of the formula $G$.
 We observe a bigger gap between BP and NN denoisers in the case of XORSAT than SAT.
This appears to be related to the fact that it is harder to train a 
(see Appendix \ref{appendix:TrainSetup} for more details.)

%
%

\subsection{Ideal target accuracy from random CSP theory}
\label{sec:Cavity}

Random CSP formulas are locally tree-like. Namely, for any fixed $r$,
the neighborhood $\Ball_G(i,r)$ centered at a uniformly random vertex $i\sim\Unif([N])$
is a tree with probability converging to one, as $N,M\to\infty$, $N/M=\alpha$.
Since BP computes correct marginals on trees,  the output of $\BPu(r)$ at  node $i$, 
is exactly the marginal of the modification of measures 
Eqs.~\eqref{eq:Tilted}, \eqref{eq:ConditionalDistr} obtained by keeping only terms in the neighborhood
$\Ball:=\Ball_{G}(i,r)$.
%
%
%

One might expect that therefore $\BPu(r)$ is asymptotically optimal among 
$r$-local denoisers.

In Figure~\ref{fig:bp_discrete_init_r2}
we compare the success rate of $\BPu(r)$ with the one of $\BPc(r)$, which differs from $\BPu(r)$ uniquely in the initialization. Recall that one BP message is associated to each edge
of the factor graph $(V,F,E)$, each message being a probability distribution over $\{+1,-1\}$
(equivalently, a likelihood ratio). 
$\BPc(u)$ initializes message to be i.i.d. from the asymptotic distribution of the root marginal
of an equivalent tree model. This distribution
be approximately sampled numerically by using a 
sampled distributional recursion method (a.k.a. `population dynamics' 
\cite{mezard2001bethe}). We refer to Appendix \ref{app:Analytical} for explicit definitions and algorithms.
%
%
%
%
%

The next theorem establishes that this `cavity' initialization is optimal (among local algorithms) with respect to the score matching 
objective of Eq.~\eqref{eq:ScoreMatching}. We refer to the appendix for background on the  local
convergence assumption: suffices to say that this is expected to hold for $\alpha\in[0,\alpha_{\smllc}(k,t))$, where  the `condensation phase transition' satisfies $\alpha_{\smllc}(k,t)\ge \alpha_{\sd}(k,t)$ \cite{ding2022proof}.
\begin{theorem}\label{thm:BPc}
For any $r\in\naturals$, $\eps>0$, $\ti\in (0,1)$, let $\hm_{\BPc(r)}(\bY_{\Ball_G(i,r)})$ be 
$\BPc(r)$ estimate of the conditional expectation at node $i\in V$, under the tilted distribution 
of Eq.~\eqref{eq:Tilted}.
If the measure $\mu^{\by,\ti}_G$ converges locally to the free boundary Gibbs measure under $N,M\to\infty$,
$N/M=\alpha$, then we have
(expectation is with respect to $G,\bx,\by$ as well as $I\sim\Unif([N])$)
\begin{align}
   \E\big\{\big(x_I-&\hm_{\BPc(r)}(\bY_{\Ball_G(I,r)})\big)^2\big\} 
    \le \inf_{\hm(\,\cdot\,)}  \E\big\{\big(x_I-\hm(\bY_{\Ball_G(I,r)})\big)^2\big\} +o_N(1)\,.\label{eq:BPc-claim}
    \nonumber
\end{align}
\end{theorem}
This asymptotic prediction is in agreement the experiments of Figure~\ref{fig:bp_discrete_init_r2}.
For comparison we also report the results obtained with a `warm-start' denoiser where, at each diffusion step,
we initialize BP with the outcome of the previous step. Notice that this is not a local denoiser (as information propagates
across longer distances over multiple diffusion steps).

\begin{figure}[t]
    \centering
    \includegraphics[width=\linewidth]{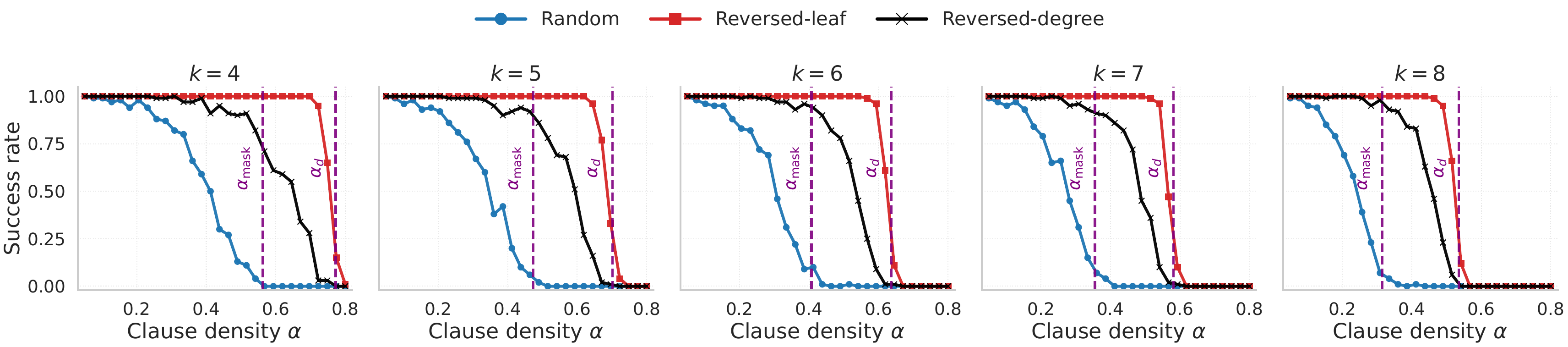}
    \caption{
        \textbf{$k$-XORSAT ($N=300$):} reversed-leaf (red) vs reversed-degree (black) vs random (blue) decoding ordering.
        Success rate as a function of clause density $\alpha$ for $k\in\{4,\dots,8\}$ obtained using discrete diffusion with BP denoiser with $r=300$. 
        The two vertical lines are the theoretical thresholds for random ($\alpha_{\mask}$) and optimal ($\alpha_{\sd}$) decoding ordering 
        at every value of $k$. Success rates are computed over 500 random formulas.
    }
    \label{fig:disc_bp}
\end{figure}



\begin{figure}[t]
    \centering
        \includegraphics[width=0.97\linewidth]{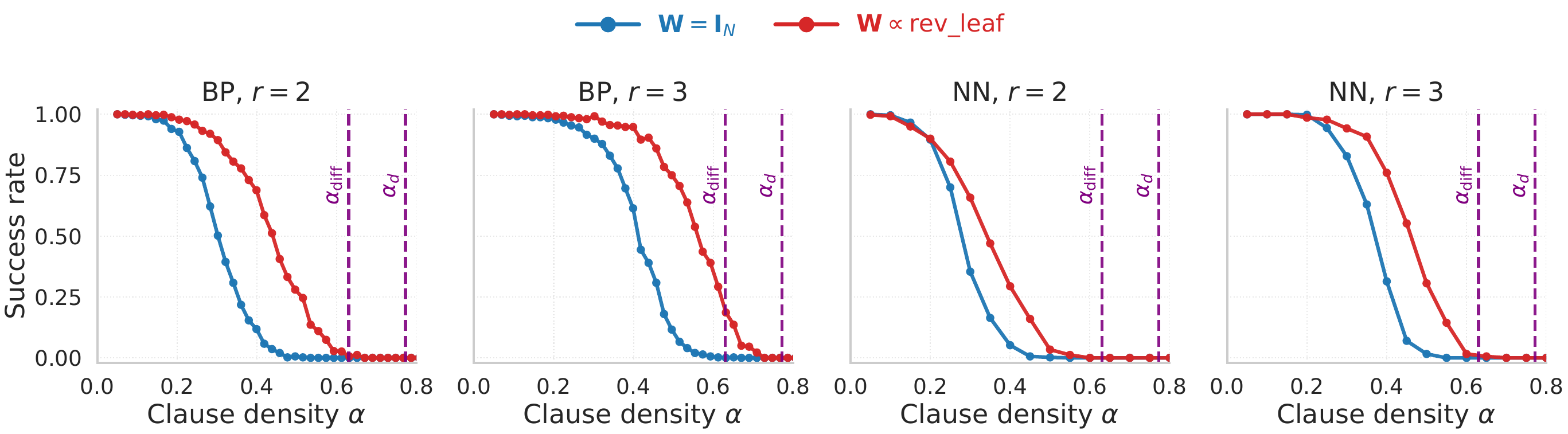}

    \caption{
        \textbf{4-XORSAT ($N=300$):} reversed-leaf (red) vs random (blue) ordering in continuous diffusion.
        Success rate as a function of clause density $\alpha$ obtained using $\bW$ in Eq.~\eqref{eq:FirstDiff},
        with a BP and NN denoiser. 
        The vertical line indicates the theoretical dynamical phase-transition threshold.
    }
    \label{fig:kxorsat_diffusion_vs_bp_reverse_leaf}
\end{figure}

\subsection{Accuracy is improved by smart ordering (continuous and discrete)}

So far we studied the behavior of denoising diffusions and discrete masked diffusions in their canonical implementation,
respectively with weight matrix $\bW=\id_N$ in Eq.~\eqref{eq:FirstDiff}, and random ordering $\ord$ in Eq.~\eqref{eq:ConditionalDistr}.
The idea that generating variables in a special order can improve the
effectiveness of  diffusion models
is not new. 
Various heuristics have been developed: the most important are based on  `model confidence'
\cite{ghazvininejad2019mask,kasai2020non} (exploiting the estimated marginals) 
or `coarse-to-fine' 
ordering  \cite{ho2022cascaded,ramesh2022hierarchical} (exploiting prior information about the variables' relations).
Despite the importance of these efforts, it is fair to say that a fundamental 
understanding of variable ordering is still missing.

We obtain a satifactory picture for discrete diffusions in XORSAT.
We begin with a negative result.
\begin{proposition}\label{propo:ConfBased}
  Consider discrete masked diffusion sampling for XORSAT. Then any ordering $\sigma$ that
  is  causal and confidence based (i.e., for each $\ell\le N$, $\sigma(\ell+1)$ only depends on on the BP marginals of $\mu^{\sigma,\ell}_G$) has the same success probability as random ordering.
\end{proposition}
We also observe that `coarse-to-fine' ordering does not have a clear interpretation for random CSPs.

In contrast, Figure~\ref{fig:disc_bp} demonstrates that a principled variable ordering  --which we refer to as `reversed leaf'--
dramatically improves over random ordering in generating XORSAT solutions via 
discrete masked diffusions. 
%
%

The reversed-leaf ordering is constructed as follows.
Given a XORSAT formula, with factor graph $G=(V,F,E)$,
recursively select one variable that appears only in one constraint (a degree-one
variable node in the factor graph), remove that variable and the corresponding constraint.
After $j$ steps, the original formula ($N$ variable with $M$ constraints) 
is reduced to a formula with $N-j$ variables and $M-j$ constraints, call it $G_j$ (whereby $G_0=G$): if the formula $G_j$ can be solved,
then any solution can also be extended to a solution of the formula $G_{j-1}$ and hence (iterating) to a solution of $G$.
The procedure can end after $M$, steps, hence resulting in a `trivial' formula with $N-M$
variables and no constraints, or halt before, if each variable appears in
two or more constraints. 

It is known that the first event occurs with high probability for 
$\alpha<\alpha_{\sd}(k)$ and the latter for $\alpha>\alpha_{\sd}(k)$ 
\cite{dubois20023,dembo2008finite}.
The reversed-leaf ordering is the reverse of the order in which 
variables are removed by the above process. This ordering is not 
unique: we assume that ties are
broken arbitrarily.

The next statement establishes that the threshold for 
discrete diffusions increases from $\alpha_{\mask}(k)$ to $\alpha_{\sd}(k)$ when replacing 
random ordering by reversed-leaf ordering.
%
\begin{theorem}\label{thm:LeafRemoval}
Let $G=(V,F,E)$ be a $k$-XORSAT instance with $|V|=N$, $|F|=M$, and $N\ge M$.
If the leaf-removal on $G$ terminates with the empty graph, 
then discrete diffusions with reversed-leaf ordering produce samples from the target distributions.

For random $k$-XORSAT, the success probability converges
to one as $N\to\infty$, $N/M=\alpha <\alpha_{\sd}(k)$.
\end{theorem}
We expect the reversed-leaf ordering to be optimal for random $k$-XORSAT (in the large $N$ limit).
We note that low degree vertices tend to be removed early in the leaf-removal process (in particular degree-one vertices can be removed before all the others).   Motivated by this remark, in Figure~\ref{fig:disc_bp} we consider a simple heuristic where we order variables by decreasing degree.
We observe that this heuristic captures a significant portion of the improvements due to reversed-leaf.

In Figure~\ref{fig:kxorsat_diffusion_vs_bp_reverse_leaf} we test whether the leaf removal order is also relevant for continuous diffusions in XORSAT. One possibility would be to use the scheme 
\eqref{eq:FirstDiff} in which $\bW$ depends on $\ell$ as to 
approximate with arbitrary precision the masked diffusion
process. Such a scheme successfully samples for all $\alpha<\alpha_{\sd}(k)$.
We try something much simpler, whereby the matrix $\bW$ is independent of $\ell$
and diagonal with $W_{ii} = 1 + c_0 (S_i - {\rm ave}(S)) / {\rm std}(S)$,
where $S_i$ is the rank of variable $i$ in the leaf removal order, and $c_0$ is a constant that we choose to be $c_0=0.55$ for BP denoiser and $c_0=0.25$ for the learned NN denoiser.  
The results of Figure~\ref{fig:kxorsat_diffusion_vs_bp_reverse_leaf} 
show that this theory-motivated approach yields a substantial improvement over 
the standard $\bW=\id_N$.

Finally, in Figs.~\ref{fig:fig1_summary_ksat} (right panel) and~\ref{fig:bp_discrete_init_r2} (two right panels), we extend these experiment to SAT. 
While no simple equivalent of the reversed-leaf ordering applies to this case,
there is a natural generalization of the reversed-degree heuristics of Figure~\ref{fig:disc_bp}.
In the case of SAT, each variable $i$ can be assigned two degrees $\deg^+(i)$, $\deg^-(i)$, 
counting the number of constraints in which it appears directed and negated. 
A highly constrained variable is a variable with large $\deg_{\min}(i)=\min(\deg^+(i), \deg^-(i))$:
we therefore generate variables in order of decreasing $\deg_{\min}$.
As shown in Figs.~\ref{fig:fig1_summary_ksat} and~\ref{fig:bp_discrete_init_r2},
the resulting procedure has a higher success rate than 
random ordering masked diffusions.

As the sampling proceeds, certain clauses are satisfied by the 
variables assigned so far and can be removed from the SAT instance. 
We denote  by $\deg_{\min}(i;\ell)$ the min-degree with respect to
such a reduced formula after $\ell$ steps. We experiment with a 
procedure that selects te next variable to be among the ones with
largest value of $\deg_{\min}(i;\ell)$.
We break the ties by sampling first the variables with the larger 
absolute value of the logit.  
We observe that this procedure (`dynamical reverse-degree') yields a 
significant increase in success rate.

\begin{figure}[t!]
  \centering
  \includegraphics[width=0.9\linewidth]{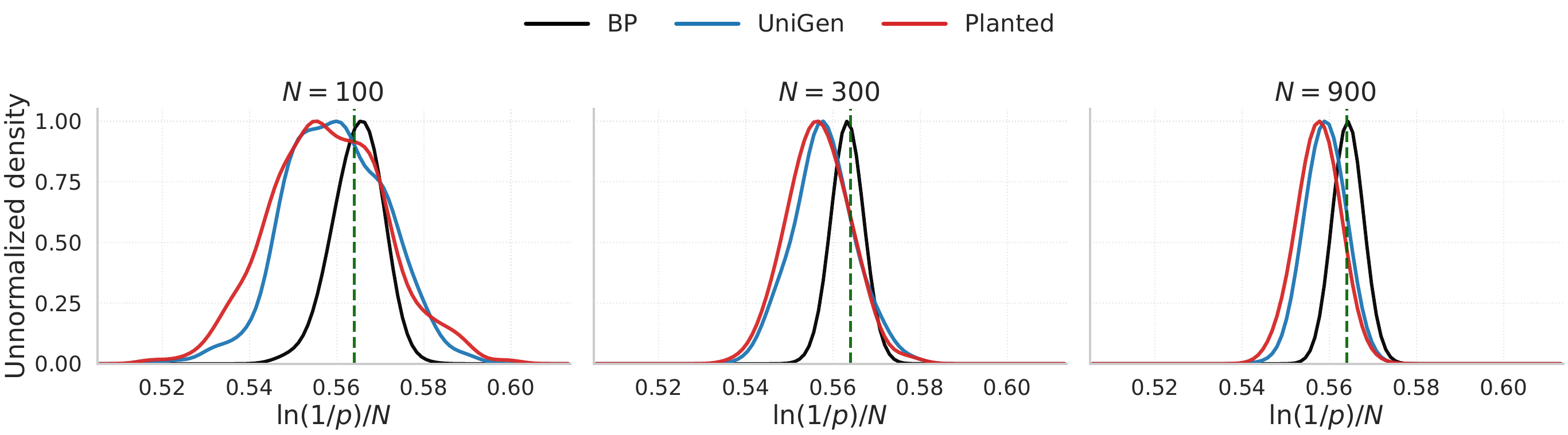}
  \caption{\textbf{4-SAT ($\alpha=2, r=3$)}  : Distribution of probabilities of sampled solutions using BP and NN (trained on either planted on UniGen formulas) discrete diffusion samplers. The verticle line corresponds to $\phi(\alpha)=\log(2)+\alpha\log(1-2^{-k})$. Averaged over 3 different formulas and 100 simulations per formula. }
  \label{fig:Prob}
\end{figure}

\subsection{Approximate uniformity test}
\label{sec:Uniformity}

Under discrete masked diffusions, let $\hmu_G^{\ord,\ell}(\,\cdot\,|\obx^{\ord,\ell})$ be the estimated conditional marginal
after sampling $\ell$ variable. It is then easy to estimate the probability of sampling $\obx$ as $-\log \hmu_G(\obx) = -\sum_{\ell=0}^{N-1} \log \hmu_G^{\ord,\ell}(\ox_{\ord(\ell+1)}|\obx^{\ord}_{1:\ell})\,$.
The expectation of this quantity over $\obx$ generated according the proposed diffusion is the entropy $H(\hmu_G)$.
Since the target distribution $\mu_G$ is uniform over solutions, their KL distance is
is $\KL(\hmu_G\|\mu_G) = \log Z_G +\E_{\obx\sim\hmu_G}\log \hmu_G(\obx)\,$. 
It is elementary to compute  $\E_G Z_G =e^{N\phi(\alpha)}$, $\phi(\alpha) := \log 2+\alpha \log(1-2^{-k})$, 
whence we get the upper bound 
\begin{align}
N^{-1}\E_G\KL(\hmu_G\|\mu_G) \le \phi(\alpha) - N^{-1}\E_G\E_{\obx\sim\hmu_G}\log \frac{1}{\hmu_G(\obx)}\,.
\end{align}

In Figure \ref{fig:Prob}, we test whether the solutions generated by our discrete diffusion algorithms
(both BP and learned) are approximately uniform in the sense of KL distance. We plot the distribution of 
$N^{-1}\log  \frac{1}{\hmu_G(\obx)}$ computed over multiple formulas and generates solutions $\obx$.
We compare this histogram with the upper bound $\phi(\alpha)$. We observe that the mean of the generated 
distributions is slightly smaller but quite close to $\phi(\alpha)$, indicating rough uniformity.
We also observe that the distribution becomes more concentrated with a mean closer to $\phi(\alpha)$
as $N$ increases, indicating increasing uniformity.
\section{Conclusion} \label{sec:discussion}

This paper
demonstrated that the uniform measure over solutions of random CSP formulas
can be a useful benchmark for generative methods.

We can summarize our findings (with supporting 
empirical and rigorous evidence):
\begin{enumerate}
\item The asymptotic theory (derived for BP denoisers)  is relevant at moderate $N,M,r$, and for learned NN denoisers
(Figs.~\ref{fig:fig1_summary_ksat}, \ref{fig:kxorsat_diffusion_vs_bp}).
\item The success rate of learned NN denoisers nearly matches the one achieved with BP denoisers (Figs.~\ref{fig:fig1_summary_ksat}, \ref{fig:kxorsat_diffusion_vs_bp}).
\item Continuous diffusions outperform discrete diffusions
(Figs.~\ref{fig:fig1_summary_ksat}, \ref{fig:kxorsat_diffusion_vs_bp}).
\item The optimal initialization for BP denoisers is not
the `uniform' one (Fig.~\ref{fig:bp_discrete_init_r2}, Theorem~\ref{thm:BPc}).
\item Smart ordering of the unmasking or denoising process leads
significant improvement in success rate
(Figs.~\ref{fig:fig1_summary_ksat}, 
\ref{fig:bp_discrete_init_r2}, \ref{fig:disc_bp}, \ref{fig:kxorsat_diffusion_vs_bp_reverse_leaf}). We achieve the 
optimal phase transition $\alpha_{\sd}(k)$ for XORSAT
(\ref{fig:disc_bp}, Theorem \ref{thm:LeafRemoval}). In contrast, `confidence-based' 
unmasking does not yield gains for XORSAT (Proposition \ref{propo:ConfBased}).
\end{enumerate}

We also emphasize that our results provides guidance as to 
which regimes and metrics can be challenging for generative methods:
$(i)$~Large formula close or beyond the dynamical phase transitions $\alpha_{\sd}(k)$ (see Fig.~\ref{fig:fig1_summary_ksat}); 
$(ii)$~CSPs with special symmetry properties such as XORSAT, which lead to difficulty in training (Fig.~\ref{fig:kxorsat_diffusion_vs_bp});
$(iii)$~Measuring uniformity of the generated solutions
(e.g., the KL bound of Fig.~\ref{fig:Prob}) can yield a very stringent 
test.

An important feature of using random CSPs as benchmark is that
theory-motivated methods such as BP provide a very strong baseline to compare to. 
\section*{Acknowledgements}
We are grateful to Neeraja Abhyankar, Zachary Frangella, Ayush Jain, Marc Laugharn, Rahul Ponnala, Sahasrajit Sarmasarkar, Andreas Santucci and Pulkit Tandon, for several
conversations about this work.

\bibliographystyle{amsalpha}

\newcommand{\etalchar}[1]{$^{#1}$}
\providecommand{\bysame}{\leavevmode\hbox to3em{\hrulefill}\thinspace}
\providecommand{\MR}{\relax\ifhmode\unskip\space\fi MR }
\providecommand{\MRhref}[2]{%
  \href{http://www.ams.org/mathscinet-getitem?mr=#1}{#2}
}
\providecommand{\href}[2]{#2}

\newpage
\appendix
\onecolumn
\section{Model architecture} \label{appendix:NNArchitrecture}

In this section, we formally define the architecture of the neural network employed to learn a denoiser from data. Consider a factor graph 
$G = (V,F,E)$ characterizing a CSP instance, and the noisy 
information about a solution of that instance,
$\bY$. In continuous denoising diffusions, $\bY$ is obtained by 
adding Gaussian noise to the true solution 
$\bx$. In masked discrete diffusions it is obtained by masking some randomly 
selected variables in $\bx$ (more details on the diffusion are provided in Appendix~\ref{appendix:NNTrainDetails}). 
The neural network for a radius $r$ denoiser makes a prediction of the true value of the $i$-th variable in the solution ${x}_{i}$ by taking into account 
the noisy values of the $r$-hop neighbours of variable $i$ in $G$. 

Since our architecture is recursive, we will first describe the
 architecture for $r=1$ and subsequently describe the construction
  of the radius-$r$ network from a radius-$r-1$ network. For simplicity, 
  we outline the architecture for $k$-SAT and describe the 
  (minor) differences in the case of $k$-XORSAT in 
  Appendix~\ref{appendix:XORSATArch}.

\subsection{Architecture for $r = 1$} \label{appendix:R1ArchitectureKSAT}

For a target variable $i$, let $\partial i \subseteq [M]$ denote the 
clauses in its $1-$neighbourhood in the factor graph, i.e.
 $\partial i := \{a \in F, (i,a) \in E\}$. Similarly, 
 $\partial a := \{i \in V, (i,a) \in E\} \subseteq [N]$. At a high level, 
 the architecture comprises of a multi-layer perceptron (MLP) that 
 serves as the denoiser, taking in a representation of the $1-$neighbourhood of the
  target variable $i$ (and possibly corruption level $(L-\ell)/L$ in continuous diffusion), 
  returning a prediction for the $i-$th variable $\in \mathbb{R}$ (continuous diffusion)
   or the distribution of the $i$-th variable (discrete diffusion). This $1$-neighbourhood 
   representation for variable $i$ is constructed by a taking a convex combination of a
    learned representation of each clause $a \in \partial i$, with the weights assigned 
    to each clause in the convex combination also learnt via a MLP. 
    
    More precisely, 
    our $r = 1$ architecture consists of the following components: 
\begin{itemize}
    \item A $\textsc{ClauseMLP} : \mathbb{R} \times 
    \{0,1\}^2 \to \mathbb{R}^d$ (continuous diffusion) or $\textsc{ClauseMLP} : \{0,1\}^5 \to \mathbb{R}^d$ (discrete diffusion); here $d$ is the embedding dimension. The $\textsc{ClauseMLP}$ constructs a representation $u_{a,j}$ for literal $j \in [k]$ in clause $a$. In each clause $a \in \partial i$ this MLP concatenates the noisy value of the literal ${Y}_{i_a(j)} \in \mathbb{R}$ (continuous) or a one-hot representation of ${Y}_{i_a(j)} \in \{-1,+1,*\}$ (${Y}_{i_a(j)} \mapsto \texttt{OneHot}({Y}_{i_a(j)}) \in \{0,1\}^3$, discrete---recall that $*$ represents masked variable) with a one-hot representation of the sign $s_{a, j}$.  
    \item For each clause $a$, the representation for the clause to the target variable $i$ $\mathrm{Emb}_i(a) \in \mathbb{R}^{kd}$ is then constructed by concatenating the representation of all $k$ literals $u_{a,j}, j \in [k]$. Importantly, in the concatenation the target literal $j^*$ corresponding to the $i$-th variable must go first---this ensures that the representation of the clause depends on the target variable $i$. $\textsc{ClauseMLP}$ comprises of a single linear layer followed by a $\texttt{ReLU}$ activation. 
    \item A $\textsc{WeightMLP}: \mathbb{R}^{kd} \to \mathbb{R}$ assigns a $s_i(a)$ to each $a \in \partial i$, so that the ``weight" of clause $a, w(a \mid i) \propto \exp(s_i(a))$ with $\sum_{a \in \partial i} w( a \mid i) = 1$. An (intermediate) representation for the $1-$ neighbourhood of variable $i$ is then given by the convex combination  $\widetilde{h}^{(1)}_i = \sum_{a \in \partial i} w( a \mid i) \mathrm{Emb}_i(a)$. $\textsc{WeightMLP}$ is a one hidden layer MLP (dimension $= d$) with $\texttt{Tanh}$ activation.
    \item A $\textsc{ProjMLP}: \mathbb{R}^{kd} \to \mathbb{R}^d$ that maps the intermediate representation $\widetilde{h}^{(1)}_i \in \mathbb{R}^{kd}$ to the final representation $h^{(1)}_i$. This is a single linear layer followed by a $\texttt{ReLU}$ activation. 
    \item A $\textsc{DenoiserMLP}: \mathbb{R}^{d} \times [0,1] \to \mathbb{R}$ (continuous) or $\textsc{DenoiserMLP}: \mathbb{R}^{d} \to \mathbb{R}^2$, which takes in $h^{(1)}_i$ and the normalized noise level $(L-\ell)/L$ (continuous) or just $h^{(1)}_i$ (discrete) to return an estimate for the $i-$th variable in the solution $\widehat{{ x}}_{\mathrm{sol},i}$ (continuous) or the logits $(\sigma_{+1,i},\sigma_{-1,i})$ for the $i-$th variable (discrete). We do not encode the noise level $(L-\ell)/L$ and provide it as a one-dimensional input to the network. $\textsc{DenoiserMLP}$ is a one hidden layer MLP (dimension $= d$) with $\texttt{ReLU}$ activation.
\end{itemize}
Throughout our experiments, we set $d = 128$. With this choice, the number of parameters in the radius$-1$ denoiser network are $\approx 130$K. A pseudocode for the forward pass in this network is provided in Algorithm~\ref{alg:RadOneDenoiserFwd}.

\begin{algorithm}[h] 
\caption{Radius-1 Learnt Denoiser (Forward Pass)}
\label{alg:RadOneDenoiserFwd}
\begin{algorithmic}
\STATE {\bfseries Input:} Factor graph $G = (V,F,E)$,
noisy assignment ${\boldsymbol Y} \in \mathbb{R}^N$ (continuous) or ${\boldsymbol Y} \in \{0,1,*\}^N$ (discrete); normalized time $(L-\ell)/L$ (for continuous)
\STATE {\bfseries Output:} Prediction for solution $\widehat{{\bf x}}_{\mathrm{sol},i}$ (continuous) or logits $(\sigma_{-1,i},\sigma_{+1,i})$ for target variable (discrete) 
\STATE \texttt{Compute clause embeddings}
\FOR{each clause $a \in \partial i$}
\STATE $j^* \gets \arg\min _{j \in [k]} {\bf 1} \{i_{a}(j) = i\}$ \hfill \texttt{// Index of literal corresponding to variable $i$} 
    \FOR{each literal $j \in [k]$}
        \STATE $v_{a,j} \gets {y}_{i_{a}(j)}$ $\in \mathbb{R}$ \hfill \texttt{// Continuous}
        \STATE $v_{a,j} \gets \texttt{OneHot}({y}_{i_{a}(j)})$ $\in \mathbb{R}^3$ \hfill \texttt{// Discrete}
        \STATE $u_{a,j} \gets \textsc{ClauseMLP}([\,v_{a,j},\, \texttt{OneHot}(s_{a,j})\,]) \in \mathbb{R}^d$
    \ENDFOR
    \STATE $\mathrm{Emb}_i(a) \gets \mathrm{concat}\big(u_{a,j^*},\ \{u_{a,j}\}_{j \in [k] \setminus \{j^*\}}\big) \in \mathbb{R}^{kd}$ \hfill \texttt{// Target literal goes first}
    \STATE $s_i(a) \gets \textsc{WeightMLP}(\mathrm{Emb}_i(a))$
\ENDFOR
\STATE \texttt{Aggregate clause embeddings, project and denoise}
\STATE $w(a \mid i) \gets \mathrm{softmax}_{a \in \partial i}\big(s_i(a)\big)$
\STATE $\widetilde{h}^{(1)}_i \gets \sum_{a \in \partial i} w(a \mid i)\,\mathrm{Emb}_i(a) \in \mathbb{R}^{kd}$ 
\STATE $h^{(1)}_i = \textsc{ProjMLP}(\widetilde{h}^{(1)}_i) \in \mathbb{R}^d$

\STATE $\widehat{{ x}}_{\mathrm{sol},i} \gets \textsc{DenoiserMLP}([h^{(1)}_i, \tau]) \in \mathbb{R}$ \qquad\qquad\qquad\qquad\qquad \texttt{// Continuous}
\RETURN{$\widehat{x}_{i}$}

\STATE $(\sigma_{-1,i},\sigma_{+1,i}) \gets \textsc{DenoiserMLP}(h^{(1)}_i) \in \mathbb{R}^2$ \qquad\qquad\qquad\qquad\qquad \texttt{// Discrete}
\RETURN{$(\sigma_{-1,i},\sigma_{+1,i})$}
\end{algorithmic}
\end{algorithm}

\subsection{Architecture for $r \ge 2$} 

\begin{figure}[t]
    \centering
    \subfloat[]{%
        \includegraphics[width=0.45\linewidth]{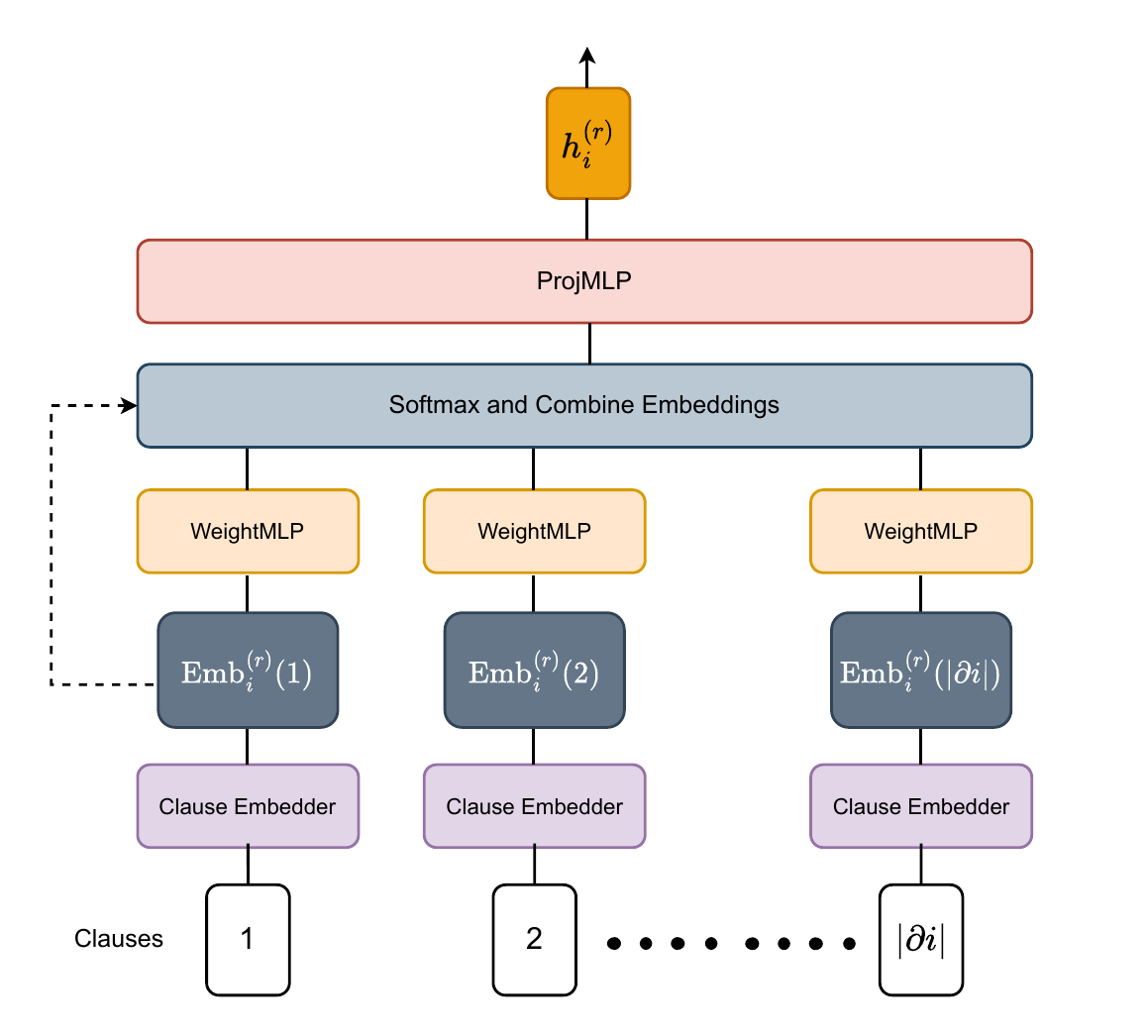}
        \label{fig:nn_architecture_1}
    }
    \hfill
    \subfloat[]{%
        \includegraphics[width=0.45\linewidth]{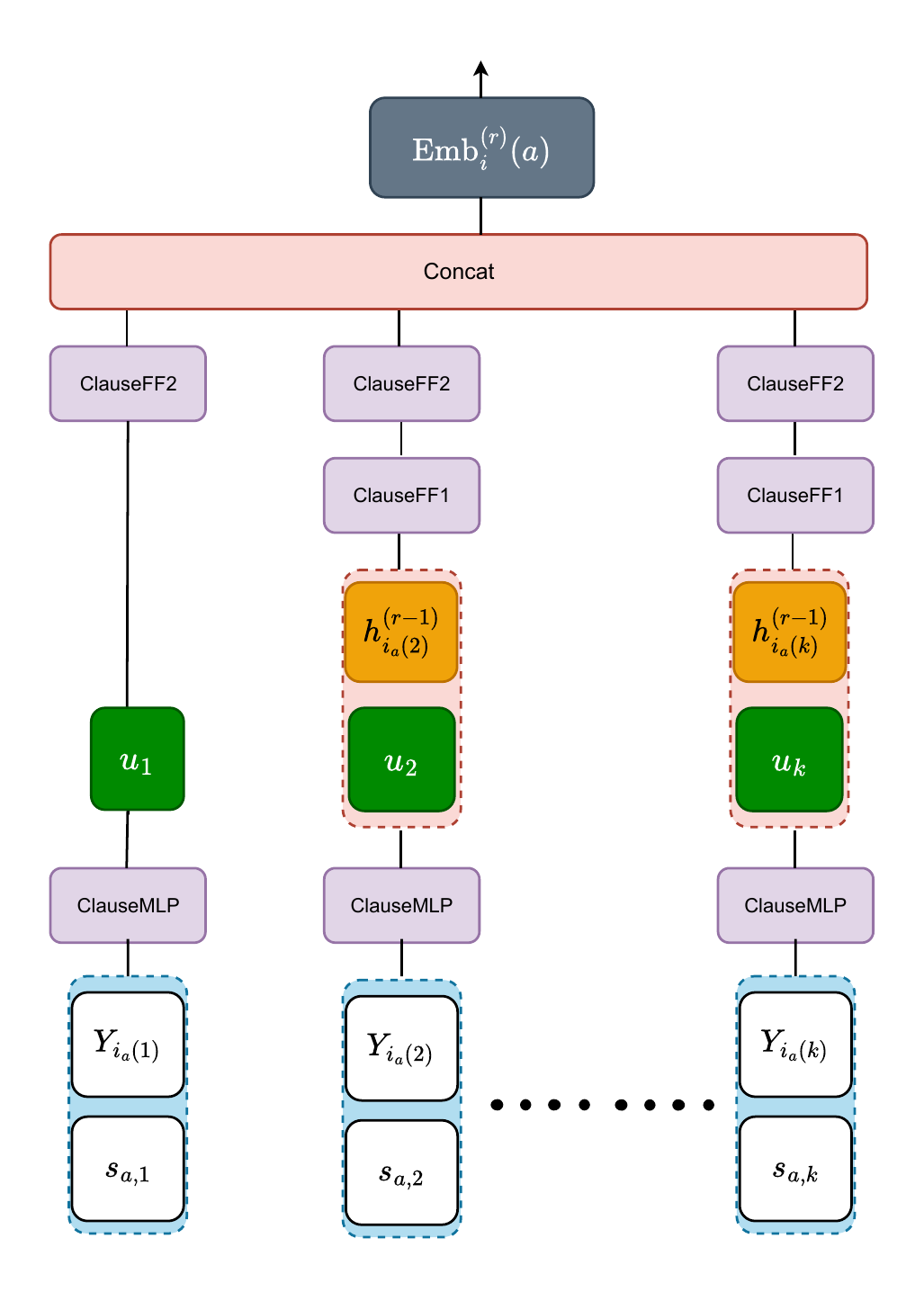}
        \label{fig:nn_architecture_2}
    }
    \caption{Neural network denoiser architecture for $r\geq2$, with target variable $i$. $(a)$ illustrates the overall architecture, assuming for simplicity that $\partial i = \{1,2,\dotsc,|\partial i|\}$. $(b)$ illustrates the detailed architecture of the $\mathrm{ClauseEmbedder}$ component (assuming again for simplicity that $i_a(1) = i$) which creates a radius $r$ representation for the clause $a$ w.r.t. variable $i$, by performing a message passing-like update that utilizes the radius $r-1$ representation of each variable present in the clause.}
    \label{fig:nn_architecture}
\end{figure}

\label{appendix:ArbitRArchitectureKSAT}
As mentioned previously, the radius $r$ embedding of variable $i$, $h^{(r)}_i$, is built from $h^{(r-1)}_1,\dotsc,h_N^{(r-1)}$ recursively; this embedding is then provided as input to the \textsc{DenoiserMLP}. The architecture is similar to the $r = 1$ case, with the primary difference being at the clause level where in addition to the signs/noisy value of each literal the $r-1$ embedding of each variable is also provided as an input to the network; thereby allowing it to incorporate a message-passing like update for radius $r$. Two additional MLPs are included: (1) $\textsc{ClauseFF1} : \mathbb{R}^{2d} \to \mathbb{R}^d$ which takes in $u_{a,j}$ (the representation of literal $j$ in clause $a$) and injects the (precomputed) radius $(r-1)$ embedding $h^{(r-1)}_{i_{a}(j)}$ of the variable corresponding to that literal $i_{a}(j)$ so as to obtain a joint representation of these in $\widetilde{u}_{a,j}$; and (2) $\textsc{ClauseFF2} : \mathbb{R}^{d} \to \mathbb{R}^d$ which further refines this embedding. Both $\textsc{ClauseFF1}$ and $\textsc{ClauseFF2}$ comprise of one linear layer followed by a $\texttt{ReLU}$ activation. 

Note that while the number of parameters of the network for $r \ge 2$ is greater than that for $r = 1$ (due to the inclusion of the \textsc{ClauseFF1} and \textsc{ClauseFF2} MLPs), the number of parameters stays constant for all $r \ge 2$. In fact, the number of parameters depends only on $k$ and $d$---with $d=128$ and $k=4$ (i.e. our experiment setting) there are $\approx 329$K parameters. 
The forward-pass pseudocode for this radius-$r$ network is provided in Algorithm~\ref{alg:RadRDenoiserFwd}; a visual depiction of the architecture is provided in Figure~\ref{fig:nn_architecture}.

\subsection{Architecture for $k-$XORSAT} \label{appendix:XORSATArch}

The architecture in the $k-$XORSAT setting is almost identical to the $k-$SAT architecture described in Appendices~\ref{appendix:R1ArchitectureKSAT} and~\ref{appendix:ArbitRArchitectureKSAT}. Recall from Section~\ref{sec:background}
that a constraint in the $k-$XORSAT problem corresponds to $ \sum_{j = 1}^k z_{i_{a}(j)} = b_{a} \text{ }\mathrm{ mod } \text{ }2$ where $x_{i_{a}(j)} = (-1)^{z_{i_{a}(j)}}$; thus, there are no signs corresponding to each literal as in the $k-$SAT setting. Correspondingly, the \textsc{ClauseMLP} is modified to take as input the bit $b_a \in \{0,1\}$ instead of $\texttt{OneHot}(s_{a,j})$ as in the $k-$SAT case. Consequently, in the continuous case $\textsc{ClauseMLP} : \mathbb{R} \times \{0,1\} \to \mathbb{R}^d$, and in the discrete case $\textsc{ClauseMLP} : \{0,1\}^4 \to \mathbb{R}^d$. All the other components including $\textsc{WeightMLP}, \textsc{ProjMLP}, \textsc{DenoiserMLP}$ and in the $r \ge 2$ case $\textsc{ClauseFF1}$ and $\textsc{ClauseFF2}$ remain the same.

\begin{algorithm}[ht]
\caption{Radius-$r$ Learnt Denoiser (Forward Pass), $r \ge 2$}
\label{alg:RadRDenoiserFwd}
\begin{algorithmic}
\STATE {\bfseries Input:} Factor graph $G = (V,F,E)$;
noisy assignment ${\boldsymbol Y} \in \mathbb{R}^N$ (continuous) or ${\boldsymbol Y} \in \{0,1,*\}^N$ (discrete);
normalized time $\tau=(L-\ell)/L$ (for continuous);
radius $r \ge 2$.
\STATE {\bfseries Output:} Prediction for solution $\widehat{{ x}}_{\mathrm{sol},i}\in\mathbb{R}$ (continuous) or logits $(\sigma_{-1,i},\sigma_{+1,i})$ for target variable (discrete).

\STATE Initialization: Construct radius-$(r-1)$ embeddings $h^{(r-1)}_{l} \in \mathbb{R}^d$ for $l \in [N]$ 

\STATE \texttt{Compute radius-$R$ clause embeddings in the $1$-neighbourhood of $i$}
\FOR{each clause $a \in \partial i$}
    \STATE $j^\star \gets \arg\min_{j \in [k]} \mathbf{1}\{i_a(j)=i\}$ \quad \texttt{// index of the literal corresponding to $i$}
    \FOR{each literal $j \in [k]$}
        \STATE $v_{a,j} \gets {y}_{i_a(j)} \in \mathbb{R}$ \hfill \texttt{// Continuous}
        \STATE $v_{a,j} \gets \texttt{OneHot}({y}_{i_a(j)}) \in \mathbb{R}^3$ \hfill \texttt{// Discrete}
        \STATE $u_{a,j} \gets \textsc{ClauseMLP}\big([\;v_{a,j},\ \texttt{OneHot}(s_{a,j})\;]\big)\in\mathbb{R}^d$
    \ENDFOR

    \STATE \texttt{Inject radius-$(r-1)$ embeddings for non-target literals}
    \FOR{each literal $j \in [k]\setminus\ j^\star $}
        \STATE $\tilde{u}_{a,j} \gets \textsc{ClauseFF1}\big([\;u_{a,j},\ h^{(r-1)}_{i_a(j)}\;]\big)\in\mathbb{R}^d$
    \ENDFOR
    \STATE $\tilde{u}_{a,j^\star} \gets u_{a,j^\star}$

    \STATE \texttt{Refine all literals}
    \FOR{each literal $j \in [k]$}
        \STATE $u^{(r)}_{a,j} \gets \textsc{ClauseFF2}\big(\tilde{u}_{a,j}\big)\in\mathbb{R}^d$
    \ENDFOR

    \STATE $\mathrm{Emb}_i^{(r)}(a) \gets \mathrm{concat}\Big(
        u^{(r)}_{a,j^\star},\ \{u^{(r)}_{a,j}\}_{j \in [k]\setminus\{j^\star\}}
    \Big)\in\mathbb{R}^{kd}$ \quad \texttt{// Target literal goes first in concatenation}
\ENDFOR
\STATE \texttt{Aggregate clause embeddings, project, and denoise}
\STATE $s^{(r)}_i(a) \gets \textsc{WeightMLP}(\mathrm{Emb}_i^{(r)}(a))\in\mathbb{R}$
\STATE $w^{(r)}(a \mid i) \gets \mathrm{softmax}_{a\in\partial i}\big(s_i^{(r)}(a)\big)$
\STATE $\widetilde{h}^{(r)}_i \gets \sum_{a \in \partial i} w^{(r)}(a \mid i)\,\mathrm{Emb}_i^{(r)}(a) \in \mathbb{R}^{kd}$
\STATE $h^{(r)}_i \gets \textsc{ProjMLP}(\widetilde{h}^{(r)}_i) \in \mathbb{R}^d$

\STATE $\widehat{{x}}_{\mathrm{sol},i} \gets \textsc{DenoiserMLP}([\,h^{(r)}_i,\ \tau\,])\in\mathbb{R}$ \hfill \texttt{// Continuous}
\RETURN{$\widehat{{x}}_{\mathrm{sol},i}$}

\STATE $(\sigma_{-1,i},\sigma_{+1,i}) \gets \textsc{DenoiserMLP}(h^{(r)}_i)\in\mathbb{R}^2$ \hfill \texttt{// Discrete}
\RETURN{$(\sigma_{-1,i},\sigma_{+1,i})$}
\end{algorithmic}
\end{algorithm}

\section{Experimental setup} \label{appendix:NNExptDetails}

In this section, we specify details pertaining to the training and 
evaluation of the learnt denoiser (both continuous and discrete) 
that were used in numerical experiments.   

\subsection{Denoiser training} \label{appendix:NNTrainDetails}
\begin{figure}[ht]
    \centering
    \includegraphics[width=0.55\linewidth]{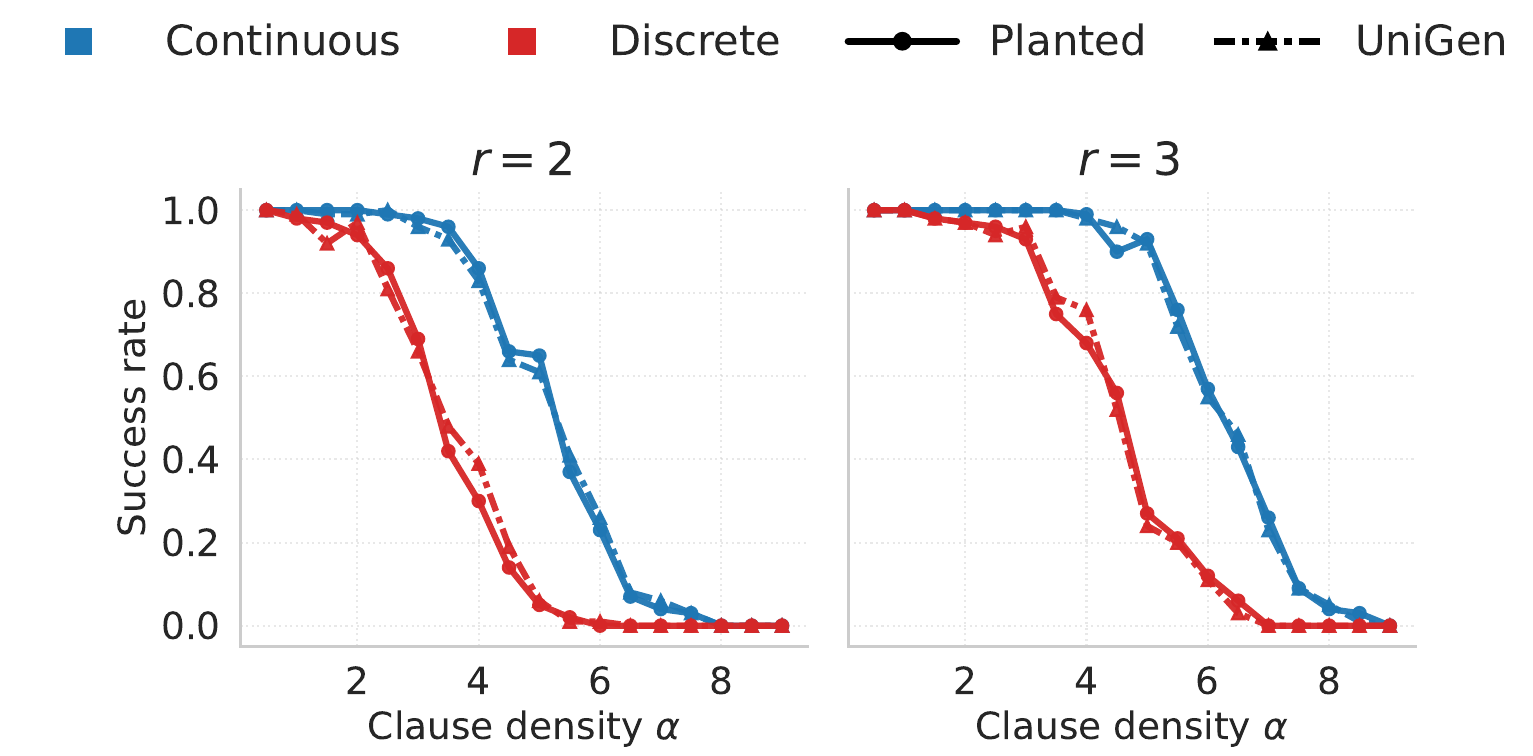}
    \caption{
        \textbf{4-SAT ($N=100$)}: Comparison of training SAT learned NN denoiser with planted formulas and solutions generated with UniGen. 
        Success rate as a function of clause density~$\alpha$ on random 4-SAT instances with $r=2$ (\textit{left}) and $r=3$ (\textit{right}). 
        Success rates are computed over 100 random formulas.
    }
    \label{fig:nn_planted_unigen}
\end{figure}

\paragraph{Dataset generation.} 
For $k$-SAT, we train the denoiser on $10{,}000$ instance--solution 
pairs  $(G, \bx)$, and for $k$-XORSAT we use $100{,}000$ such pairs, 
in both cases with $N=50$ variables and $k=4$. For $4$-SAT, 
the training data are uniformly distributed across $18$ constraint 
densities $\alpha \coloneqq M/N$ spanning $[0.5, 9]$. For $4$-XORSAT, 
the training data are uniformly distributed across $16$ values of 
$\alpha$ spanning $[0.05, 0.8]$.
In the case of $k-$SAT, we experimented with two methods of generating the training data: 
\begin{itemize}
    \item {\bf Randomly generated formula with solver generated solution.}
    In this method, a formula/factor graph $G$ was generated uniformly at random, 
    and then a solver is used to generate the corresponding solution 
    $\bx$. In order to form $G$, clauses were constructed by sampling the
     variables 
    in each clause $S_a \sim \Unif \binom{[N]}{k}$, $S_a=\{i_a(1),\dotsc,i_a(k)\}$,
    i.i.d. for $a \in [M]$ 
    (i.e. sampling a $k-$sized subset of $[N]$ uniformly at random) and then the signs of
     each literal in the clause $s_{a,j} \in \Unif\{+1,-1\}$ were assigned
      uniformly at random for $a \in [M], j \in [k]$. Upon generating $G$, 
      the solver $\mathsf{UniGen}$~\citep{Chakraborty--Meel--Vardi14, Chakraborty--Fremont--Meel--Seshia--Vardi15, Soos--Gocht--Meel20} 
      was used to generate a solution to $G$. 
       While $\mathsf{UniGen}$ does not exactly sample from the uniform measure over 
       solutions $\mu_G$, it enjoys guarantees of approximate uniformity as shown 
       in the cited literature. 
       Any randomly generated formula for which a solution was not found was discarded 
       from the training data set. 
    \item {\bf Random assignment and planted formula.} In this case, 
    first a solution is sampled $\bx\sim \Unif(\{-1,+1\}^{N})$, and then 
    a random CSP is sampled as $G \sim \Unif\{\widetilde{G}: \bx \text{ is a solution to }
    \widetilde{G}\}$. For the latter, variables in each clause are sampled
     $S_a \sim \Unif\binom{[N]}{k}$ i.i.d. for $a \in [M]$  and then 
     the signs of each literal in the clause $(s_{a,1},\dotsc,s_{a,k}) \in 
     \Unif( \mathscr{S}_a)$ where $\mathscr{S}_a \subset \{+1,-1\}^k, 
     |\mathscr{S}_a| = 2^k-1$ consists of all possible sign assignments except the 
     one where $s_{a,j} = - x_{i_a(j)}$ for all $j \in [k]$ 
     (i.e., the sign assignment for which none of the $k$ literals would evaluate to 
     $\mathrm{True}$). We note that this method of generating training data 
     induces a distribution mismatch since at test-time we always evaluate our 
     solver on a randomly generated formula (see Appendix~\ref{appendix:NNEvals} 
     for details).  We observe (see Figure~\ref{fig:nn_planted_unigen}) that results
      from models trained with planted training data are very similar 
      to results from models trained with $\mathsf{UniGen}$  generated data. 
      Unless mentioned otherwise we train models on data generated with planted solutions. 
\end{itemize}

For $k$-XORSAT, we construct the dataset by sampling a random instance 
( $S_a=\{i_a(1),\dotsc,i_a(k)\}\sim \Unif\binom{[N]}{k}$ and 
$b_a \sim \Unif(\{-1,+1\})$ for $a \in [M]$ i.i.d.) and solving the 
resulting system of linear equations via Gaussian elimination. 
For an underspecified system, Gaussian elimination will yield several 
``free" variables that can be set to either of $+1$ or $-1$ 
to provide a  solution to the system. Depending on how we set these
 free variables we obtain two methods to construct the dataset---one with
  `biased solutions,' where the free variables are set to $+1$ and one 
  with `unbiased solutions,' where the free variables are sampled
   $\sim \Unif\{-1,+1\}$. 

\begin{figure}[ht]
    \centering
    \includegraphics[width=0.55\linewidth]{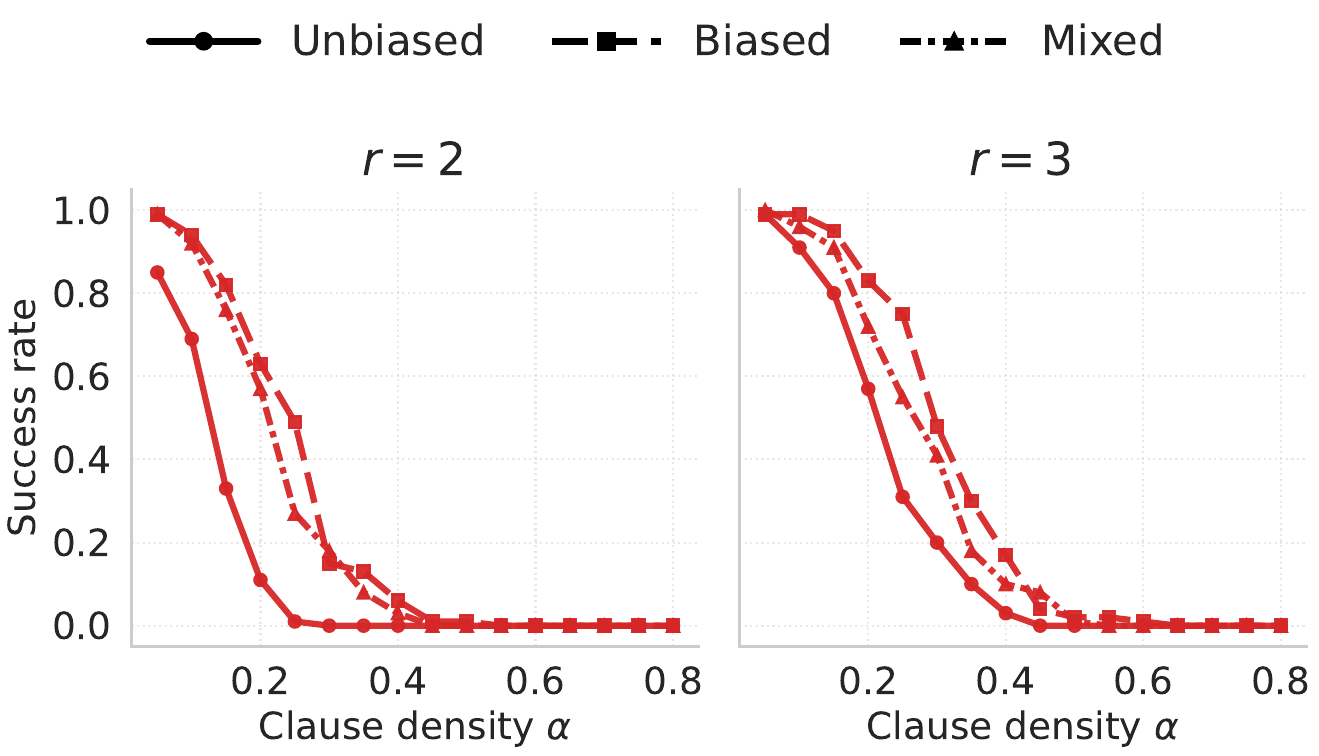}
    \caption{
    \textbf{4-XORSAT ($N=100$):} Comparison of learned NN denoisers trained on biased solutions, unbiased solutions, and a mixed curriculum that begins with biased solutions and then transitions to unbiased solutions. Success rate as a function of clause density~$\alpha$ on random 4-XORSAT instances with $r=2$ (\textit{left}) and $r=3$ (\textit{right}). Success rates are computed over 100 random formulas.       
    }
    \label{fig:nn_bias_unbiased}
\end{figure}

It is easy to see that unbiased solutions are uniformly random solutions by 
construction.
 However, in the discrete masked diffusions, we observe that denoisers trained from
 scratch on unbiased solutions perform poorly. We hypothesize that 
 this is due to vanishing (or near-zero) gradients early in training. 
 In order to overcome this difficulty, we adopt a two-stage procedure: 
 we first train on biased solutions 
 for the initial $50\%$ of epochs and then fine-tune on 
 unbiased solutions (Figure~\ref{fig:nn_bias_unbiased}). We do not observe 
 this degradation for continuous diffusions and for larger $r$ in 
 the discrete diffusions, and in those cases we train directly on
  unbiased solutions.

\begin{figure}[ht]
    \centering
    \includegraphics[width=0.55\linewidth]{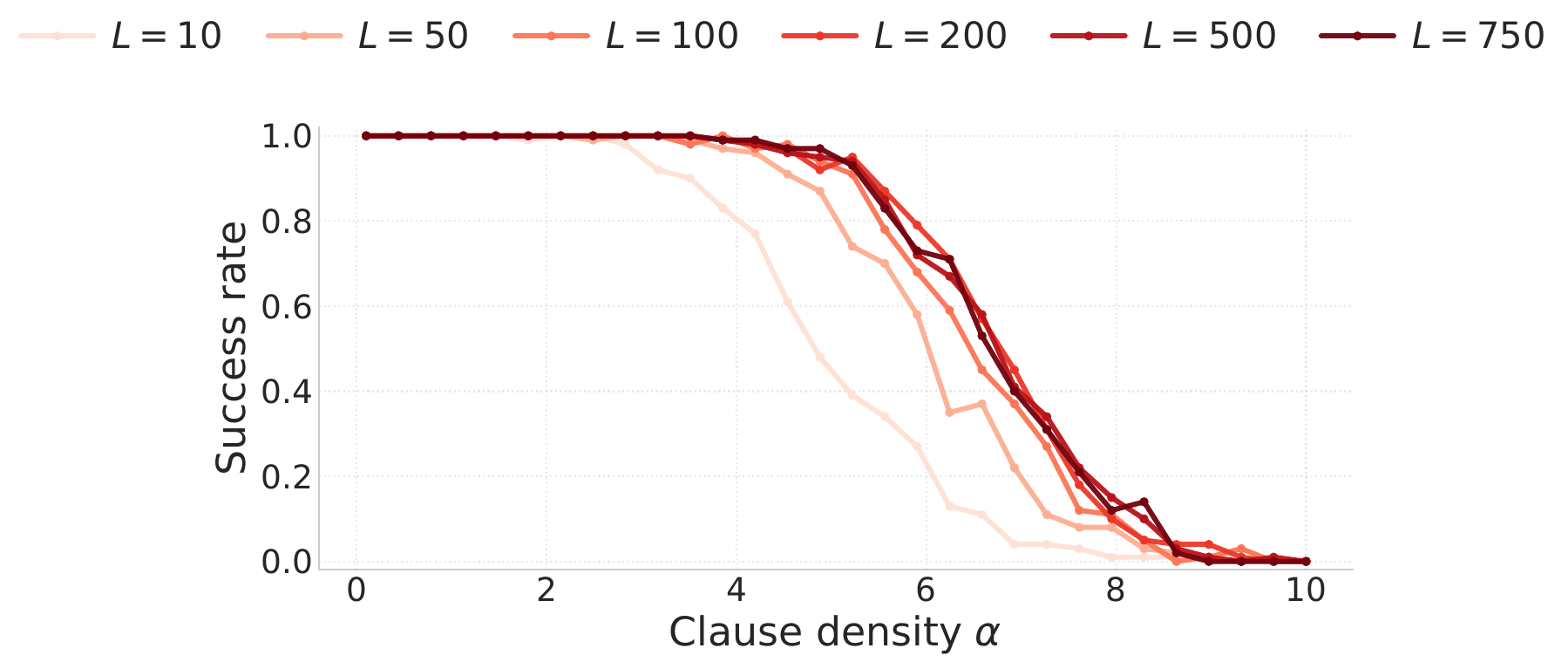}
    \caption{
        \textbf{4-SAT ($N=100$)}:
        Comparison of diffusion timesteps ($L$) for continuous diffusion with a local BP denoiser. Success rate as a function of clause density $\alpha$ on 100 random instances, evaluated with $r=3$.
    }
    \label{fig:cont_bp_timesteps}
\end{figure}

\paragraph{Training details.} \label{appendix:TrainSetup}
For $k$-SAT, we train continuous diffusion denoisers for $100$ epochs and discrete-diffusion denoisers for $200$ epochs. For $k$-XORSAT, we train both continuous and discrete denoisers for $200$ epochs, except that continuous-diffusion denoisers at $r\in{6,9}$ are trained for $300$ epochs. For the continuous NN denoiser, $L$ was chosen to be $500$ and a cosine schedule was used~\citep{Nichol--Dhariwal21} to generate diffusion parameters $\gamma_{\ell}, \ti_{\ell}$. A simplified pseudocode (with batch size $1$) for the training of both the continuous learnt diffusion denoiser (DDPM) and discrete learnt diffusion denoiser (MDM) is provided in Algorithm~\ref{alg:trainingNNDenoiser}. In the implementation, all $k$-SAT models were trained with batch size $32$ using AdamW ($\beta_1=0.9$, $\beta_2=0.999$). Because $k$-XORSAT is evaluated at lower clause densities, which yields fewer effective message-passing interactions per instance, we use a larger training set ($100{,}000$ CSP instance--solution pairs) and a correspondingly larger batch size of $256$. For both, we use a learning rate of $3\times 10^{-4}$ with linear warm-up followed by a cosine decay schedule. Following earlier works on diffusion models~\citep{anil2025interleaved, song2021score}, we use exponential moving average (EMA) of parameters with decay $0.99$ for evaluation. Unless stated otherwise, we do not perform hyperparameter tuning. 
During model training, we periodically log the performance (i.e. success rate) on a held-out validation set at the median $\alpha$ ($0.45$ for $k$-XORSAT and $5$ for $k$-SAT). We save a new model checkpoint everytime either $(i)$ the validation performance exceeds the best previously obtained validation performance, or $(ii)$ the training loss drops more than $2\%$ from the last epoch. We report our test performance (across all $\alpha$ values) with the latest saved model. 

\paragraph{Diffusion timesteps.}
To select an appropriate number of diffusion timesteps ($L$), Figure~\ref{fig:cont_bp_timesteps} compares the performance of continuous diffusion with a BP denoiser on 4-SAT across different values of $L$. We observe that performance saturates by $L\approx500$, with no additional gains for larger $L$. Accordingly, we use $L=500$ timesteps for continuous diffusion in all experiments, including those with learned NN denoisers.

\begin{algorithm}[ht]
\caption{Training for Learnt Denoisers (Continuous DDPM and Discrete MDM)}
\label{alg:trainingNNDenoiser}
\begin{algorithmic}
\STATE {\bf Input:} Training set $\mathcal{D}=\{(G_j, {\boldsymbol x}_{\mathrm{sol},j})\}_{j = 1}^{n_{\mathrm{train}}}$; epochs $E$; radius $r$; $L$; Noising schedule $\ti_{\ell}, \ell \in [L]$ (for continuous, recall~\eqref{eq:FirstDiff}). 
\STATE {\bf Models:} $\textsc{ContRadiusRNet}_\theta$ (continuous), $\textsc{DiscRadiusRNet}_\phi$ (discrete).
\STATE {\bf Output:} Trained parameters $\theta$ and $\phi$.

\FOR{epoch $=1,\dotsc,E$}
    \FOR{each training sample $(G, {\boldsymbol x}_{\mathrm{sol}})\in\mathcal{D}$}

        \STATE \texttt{Continuous corruption}
        \STATE  Sample $\ell \sim \mathsf{Unif}([L])$ 

        \STATE Sample ${\boldsymbol \epsilon} \sim\mathcal{N}(0,\id_N)$ and form noisy solution ${\boldsymbol Y}_{\ell} = \sqrt{\ti_{\ell}}\,{\boldsymbol x}_{\mathrm{sol}} + \sqrt{1-\ti}_{\ell} \,\epsilon \in \mathbb{R}^N$ 

        \STATE \texttt{Discrete corruption}
        \STATE Sample corruption level $\ell \sim \mathsf{Unif}([N])$ 
        \STATE Sample a subset $S \subseteq [N]$ with $|S| = N-\ell$ uniformly at random
        \STATE ${\boldsymbol Y}_{t, S} \gets *$ for $k \in S$ (masking), and ${\boldsymbol Y}_{t, [N] \setminus S} \gets {\boldsymbol x}_{\mathrm{sol}, [N] \setminus S}$  

        \STATE \texttt{Forward pass}
        \STATE $\widehat{{\boldsymbol x}}_{\mathrm{sol}} \gets \textsc{ContRadiusRNet}_\theta.\textsc{fwd}(G,{\boldsymbol Y}_{\ell},r,(L-\ell)/L)\in\mathbb{R}^{N}$ \hfill \texttt{// Continuous, recover entire vector}
        \STATE Sample target variable $i \sim \mathrm{Unif}([N])$
        \STATE $(\sigma_{-1,i}, \sigma_{+1,i}) \gets \textsc{DiscRadiusRNet}_\phi.\textsc{fwd}(G,{\boldsymbol Y}_{\ell},r,i) \in\mathbb{R}^{2}$. \hfill \texttt{// Discrete, recover logits for target variable}

        \STATE \texttt{Loss computation}
         \STATE $\mathcal{L}_{\mathrm{cont}} \gets \|\boldsymbol{{\widehat{x}}}_{\mathrm{sol}} - {\boldsymbol x}_{\mathrm{sol}}\|_2^2 = \textsc{MSE}(\boldsymbol{ \widehat{x}}_{\mathrm{sol}}, {\boldsymbol x}_{\mathrm{sol}})$ . \hfill \texttt{// Continuous}
        \STATE $\mathcal{L}_{\mathrm{disc}} \gets -\log\left(\frac{\exp(\sigma_{{x}_{\mathrm{sol},i},\,i})}{\exp(\sigma_{-1,i})+\exp(\sigma_{1,i})}\right)
        \;=\; \textsc{CrossEntropy}\big((\sigma_{-1,i},\sigma_{1,i}),\, { x}_{\mathrm{sol},i}\big)$. \hfill \texttt{// Discrete}

        \STATE \texttt{Backward pass}
        \STATE $\theta \gets \theta - \eta\,\nabla_\theta \mathcal{L}_{\mathrm{cont}}$. \hfill \texttt{// Continuous}
        \STATE $\phi \gets \phi - \eta\,\nabla_\phi \mathcal{L}_{\mathrm{disc}}$. \hfill \texttt{// Discrete}

    \ENDFOR
\ENDFOR
\end{algorithmic}
\end{algorithm}

\subsection{Diffusion and inference} \label{appendix:NNEvals}

\paragraph{Diffusion.}
Upon obtaining trained models \textsc{ContRadiusRNet}$_\theta$ and $\textsc{DiscRadiusRNet}_{\phi}$ we utilize them to generate a 
solution to an arbitrary instance $G$. The pseudocode for this diffusion process is provided in Algorithm~\ref{alg:ContDiffNNDenoiser} (continuous diffusion) and Algorithm~\ref{alg:DiscreteDiffNNDenoiser} (discrete diffusion). Recalling our time parametrization in continuous diffusion from~\eqref{eq:FirstDiff}, we note that a smaller $\ell$ corresponds to higher noise.

\begin{algorithm}[ht]
\caption{Continuous Diffusion Sampling of CSP Solutions with NN Denoiser}
\label{alg:ContDiffNNDenoiser}
\begin{algorithmic}
\STATE {\bfseries Input:} CSP $G$, Diffusion steps $L$; noise schedule $\{\beta_{\ell},\ti_{\ell}\}_{\ell=1}^L$ with $\ti_{\ell}=\prod_{i=\ell}^{L-1}(1-\beta_i)$;
trained denoiser $\textsc{ContRadiusRNet}_{\theta}$
\STATE {\bfseries Output:} Candidate solution $\widehat{\boldsymbol x}\in\{-1,+1\}^N$.

\STATE Sample ${\boldsymbol Y}_0 \sim \mathcal{N}({\boldsymbol 0}, \id_N)$.
\FOR{$\ell = 0, 1, \dotsc, L-1$}
    \STATE $\widehat{\boldsymbol x}_{\mathrm{sol}} \gets \textsc{ContRadiusRNet}_{\theta}(G, {\boldsymbol Y}_{\ell},(L-\ell)/L)\in\mathbb{R}^N$. \hfill \texttt{// Denoise }
    \STATE $\widehat{\boldsymbol{\epsilon}} \gets \dfrac{{\boldsymbol Y}_{\ell}- \sqrt{\ti_{\ell}}\,\widehat{\boldsymbol x}_{\mathrm{sol}}}{\sqrt{1-\ti_{\ell}}}\in\mathbb{R}^N$. \hfill \texttt{// Implied noise}
    \STATE $\boldsymbol{\mu}_{\ell} \gets \dfrac{1}{\sqrt{1-\beta_{\ell}}}\left({\boldsymbol Y}_{\ell} - \dfrac{\beta_{\ell}}{\sqrt{1-\ti_{\ell}}}\,\widehat{\boldsymbol{\epsilon}}\right)\in\mathbb{R}^N$. \hfill \texttt{// DDPM posterior mean}
    \STATE $\sigma_{\ell}^2 \gets \beta_{\ell}$
    \STATE Sample ${\boldsymbol z}\sim\mathcal{N}({\boldsymbol 0},\id_n)$
    \STATE ${\boldsymbol Y}_{\ell+1} \gets \boldsymbol{\mu}_{\ell} + \sigma_{\ell}{\boldsymbol z}$.
\ENDFOR
\STATE $\widehat{\boldsymbol x} \gets \mathrm{sign}({\boldsymbol Y}_L) \in \{-1,+1\}^N$ 
\RETURN{$\widehat{\boldsymbol x}$}
\end{algorithmic}
\end{algorithm}

\begin{algorithm}[h]
\caption{Discrete Diffusion Sampling of CSP Solutions with NN Denoiser}
\label{alg:DiscreteDiffNNDenoiser}
\begin{algorithmic}
\STATE {\bfseries Input:} CSP $G$; Trained denoiser $\textsc{DiscRadiusRNet}_{\phi}$.
\STATE {\bfseries Output:} Candidate solution $\widehat{\boldsymbol x}\in\{-1,+1\}^n$.

\STATE ${\boldsymbol Y} \gets \{*\}^n$, $S_{\mathrm{masked}} = [n]$
\WHILE{$|S_{\mathrm{masked}}| > 0$}
    \STATE Sample $i \sim \mathrm{Unif}(S_{\mathrm{masked}})$
    \STATE $(\sigma_{-1,i},\sigma_{+1,i}) = \textsc{DiscRadiusRNet}_{\phi}(G, {\boldsymbol Y}, i)$    \hfill \texttt{ // Probability assignment for target}
    \STATE Sample $b \sim \mathsf{Bern}\left(\frac{\exp(\sigma_{+1,i})}{\exp(\sigma_{+1,i})+\exp(\sigma_{-1,i})}\right)$
    \STATE $Y_i \gets 2b-1$ \hfill{\texttt{// Update } ${\boldsymbol Y}$ \texttt{ to reveal one more variable}}
    \STATE $S_{\mathrm{masked}} \gets S_{\mathrm{masked}} \setminus \{i\}$
\ENDWHILE
\STATE $\widehat{{\boldsymbol x}} \gets {\boldsymbol Y}$
\RETURN{$\widehat{\bf x}$}
\end{algorithmic}
\end{algorithm}

\paragraph{Evaluation.}
For each experimental setting and each value of $\alpha$, we evaluate on $500$ randomly generated test formulae and report the fraction for which the trained learnt solver returns an accurate solution. We emphasize that regardless of how the training data is generated (random formula or planted solution), at test-time the \emph{formula} is generated uniformly at random. Unlike during training, we do not discard unsatisfiable CSP instances, and count an unsatisfiable instance as a failure for the solver. 
Because the model architecture does not depend on $N$, we use the same model checkpoints trained at $N=50$ when evaluating larger values of $N$. Similarly, we use a single checkpoint trained on the mixture over $\alpha$ and report performance separately on test sets at each fixed $\alpha$. 
We trained models for $r=2,3,6$ and $9$.

\subsection{Compute requirements} 
All experiments were run on a single node with 128 CPU cores and NVIDIA A100 80\,GB GPUs, using Ubuntu 22.04 and CUDA 12.1. Evaluations of the BP denoiser require no training and were performed entirely on CPU. For learned denoiser evaluations, we trained eight models each for continuous and discrete diffusions: $k$-SAT with $r\in\{2,3,6,9\}$ and $k$-XORSAT with $r\in\{2,3,6,9\}$ ($N = 50$ was fixed throughout). At inference time, we reuse the corresponding checkpoint while varying $\alpha$, $N$, and (for $k$-XORSAT) the variable ordering. Each of the sixteen training runs took less than two GPU-hours.

\section{Details on belief propagation} \label{appendix:BPDetails}

In this section we describe the belief propagation (BP) 
message passing algorithm that we use as a \emph{benchmark denoiser} 
both for \emph{discrete} and \emph{continuous} diffusion samplers.  
We refer to \cite{MezardMontanari} for a general introduction to
belief propagation for constraint satisfaction problems,
and to \cite{montanari2007solving,ghio2024sampling} 
for earlier applications to diffusions sampling.

\subsection{BP denoiser for $k$-SAT discrete diffusion}

\label{appendix:BP_discrete}

\paragraph{Random $k$-SAT instances}
As described in the main text, a $k$-SAT clause 
$a\in[M]$ takes the form
\begin{equation}
\bigl(z_{i_a(1)}\bigr)\ \vee\ \bigl(z_{i_a(2)}\bigr)\ \vee\ \cdots\ \vee\ \bigl(z_{i_a(k)}\bigr)=+1,
\qquad
z_{i_a(j)} \;=\; s_{a,j}\,x_{i_a(j)},
\label{eq:ksat_clause_paper_notation}
\end{equation}
where $i_a(1),\dots,i_a(k)\subseteq[N]$
 are distinct indices, $s_{a,j}\in\{+1,-1\}$
 are clause signs, and $x\vee x'=\max(x,x')$ 
 corresponds to logical OR under the $\{+1,-1\}$ 
 encoding. For $i\in\partial a$, we write $s_{a,i}:=s_{a,j}$ whenever $i=i_a(j)$, and
$\bar s_{a,i}:=-\,s_{a,i}$.
Thus $x_i=s_{a,i}$ satisfies clause $a$ through the literal attached to edge
$(a,i)$, while $x_i=\bar s_{a,i}$ does not.

\paragraph{Factor graph representation.}
We represent an instance by a bipartite factor graph with variable nodes $V=\{1,\dots,N\}$ and clause (factor) nodes $F=\{1,\dots,M\}$.
We write $\partial a\subseteq V$ for the set of variables appearing in clause $a$, and $\partial i\subseteq F$ for the set of clauses incident to variable $i$.
Whenever $i\in\partial a$ we denote by $\partial a\setminus i$ and $\partial i\setminus a$ the corresponding neighborhoods with the edge $(a,i)$ removed.


For an adjacent variable node $i$ and clause node $a$ (i.e.\ $i\in\partial a$), we define the two subsets of neighboring clauses of $i$ (excluding $a$):
\begin{equation}
\partial_{+} i(a)
\;\equiv\;
\bigl\{\, b\in\partial i\setminus\{a\}\ :\ \bar s_{b,i}=\bar s_{a,i}\,\bigr\},
\qquad
\partial_{-} i(a)
\;\equiv\;
\bigl\{\, b\in\partial i\setminus\{a\}\ :\ \bar s_{b,i}=-\bar s_{a,i}\,\bigr\}.
\label{eq:paper_neighbor_sets}
\end{equation}
Intuitively, $\partial_{+} i(a)$ are clauses whose preferred value for $x_i$ agrees with clause $a$, while $\partial_{-} i(a)$ are those that disagree.

We next define the BP messages on the directed edges $i\to a$ and $a\to i$ and give their update equations; we then explain how these same equations are used as a denoising step in our discrete and continuous diffusion samplers.

\paragraph{Variable-to-clause update (iterative form).}
Let $r\in\{0,1,2,\dots\}$ index BP iterations, and denote by
$u^{(r)}_{b\to i}$ and $h^{(r)}_{i\to a}$ the clause-to-variable and
variable-to-clause messages at round~$r$. Given the incoming clause-to-variable
messages $\{u^{(r)}_{b\to i}\}_{b\in\partial i\setminus a}$, the cavity field
$h^{(r+1)}_{i\to a}$ is updated by splitting neighbors according to~\eqref{eq:paper_neighbor_sets}:
\begin{equation}
h^{(r+1)}_{i\to a}
\;=\;
\sum_{b\in \partial_{+} i(a)} u^{(r)}_{b\to i}
\;-\;
\sum_{b\in \partial_{-} i(a)} u^{(r)}_{b\to i}.
\label{eq:paper_var_to_clause_iter}
\end{equation}
This is the ``sum of incoming influences with a sign'' rule: clauses in
$\partial_{+} i(a)$ contribute positively, while clauses in $\partial_{-} i(a)$
contribute negatively.

\paragraph{Clause-to-variable update (iterative form).}
For a clause $a$ and a variable $i\in\partial a$, the message
$u^{(r+1)}_{a\to i}$ is updated as a function of the incoming cavity fields
$\{h^{(r+1)}_{j\to a}\}_{j\in\partial a\setminus i}$ from the other variables
in the clause:
\begin{equation}
u^{(r+1)}_{a\to i}
\;=\;
-\frac12 \log\!\left(
1 - (1-\epsilon)\,2^{-(k-1)}\,
\prod_{j\in\partial a\setminus i}\bigl(1-\tanh\!\bigl(\bar s_{a,j}\,h^{(r+1)}_{j\to a})\bigr)\right),
\label{eq:paper_clause_to_var_iter}
\end{equation}
where $\bar s_{a,j}\in\{\pm1\}$ is the unsatisfying value of $x_j$ for clause
$a$, and $\epsilon\in[0,1)$ is an optional noise/relaxation parameter used to
avoid hard warnings.\footnote{Setting $\epsilon=0$ gives the standard BP update;
$\epsilon>0$ weakens clause influences.}

\medskip
Equations~\eqref{eq:paper_var_to_clause_iter}--\eqref{eq:paper_clause_to_var_iter}
define one synchronous BP iteration: first compute all
$\{h^{(r+1)}_{i\to a}\}$ from $\{u^{(r)}_{b\to i}\}$, then compute all
$\{u^{(r+1)}_{a\to i}\}$ from $\{h^{(r+1)}_{j\to a}\}$, and repeat for
$r$ rounds.

\paragraph{From messages to marginals.}
At any BP round~$r$, once the clause-to-variable messages $\{u^{(r)}_{a\to i}\}$ have been updated,
we form an (approximate) variable ``field'' by summing incoming influences with their signs:
\begin{equation}
H_i^{(r)}
\;=\;
\sum_{a\in\partial i} s_{a,i}\,u^{(r)}_{a\to i},
\label{eq:paper_total_field}
\end{equation}
and the corresponding magnetization and Bernoulli marginal are
\begin{equation}
m_i^{(r)}=\tanh\!\bigl(H_i^{(r)}\bigr),
\qquad
\widehat{\mathbb{P}}^{(r)}(x_i=+1)=\frac{1+m_i^{(r)}}{2}.
\label{eq:paper_marginal_from_field}
\end{equation}

\paragraph{Initialization and stopping.}
We initialize the clause-to-variable messages $\{u^{(0)}_{a\to i}\}$ according to the selected policy:
(i) \emph{zero init}, $u^{(0)}_{a\to i}=0$ for all directed edges; (ii) \emph{warm init}, where we reuse the previous-step messages on the surviving edges from the previous diffusion step $t$; or (iii) \emph{cavity init}, where $u^{(0)}_{a\to i}$ are sampled i.i.d.\ from the population-dynamics approximation of the fixed-point distribution $\widehat Q(u)$ (see Appendix~\ref{appendix:CavityDetails} ).
Starting from $\{u^{(0)}_{a\to i}\}$, we run $r$ synchronous BP rounds using
\eqref{eq:paper_var_to_clause_iter}--\eqref{eq:paper_clause_to_var_iter}.
Optionally, we terminate early if the maximum change in messages across one round falls below a tolerance,
\begin{equation}
\max_{(a,i)\in E}\Bigl|u^{(r+1)}_{a\to i}-u^{(r)}_{a\to i}\Bigr| \;<\; \tau,
\label{eq:paper_early_stop}
\end{equation}
but in our sweeps we primarily compare methods at a fixed computational budget by fixing $r$ and setting $\tau=0$.

\paragraph{Implementation details.} Our implementation is mathematically equivalent to Eqs.~\eqref{eq:paper_var_to_clause_iter}, \eqref{eq:paper_clause_to_var_iter},  but is organized to avoid storing both message families explicitly. We maintain only the clause-to-variable messages $\{u_{a\to i}\}$ and reconstruct the variable-to-clause cavity fields $\{h_{i\to a}\}$ on the fly via the cavity identity. Specifically, defining the full field at variable $i$ as
\begin{equation}
H_i \;=\; \sum_{b\in\partial i} s_{b,i}\,u_{b\to i},
\label{eq:impl_field}
\end{equation}
the cavity field for the directed edge $(i\to a)$ is obtained exactly by removing the contribution of the destination clause,
\begin{equation}
h_{i\to a}
\;=\;
\sum_{b\in\partial i\setminus\{a\}} s_{b,i}u_{b\to i}
\;=\;
H_i - s_{a,i}u_{a\to i}.
\label{eq:impl_cavity_identity}
\end{equation}
This allows us to evaluate the clause update \eqref{eq:paper_clause_to_var_iter} without a separate state for $h$. For each clause $a$ and neighbor $i\in\partial a$, the update depends on a leave-one-out product over the remaining variables $j\in\partial a\setminus\{i\}$; in code we compute the oriented terms
\(
t_{j\to a} \equiv 1-\tanh\!\bigl(s_{a,j}\,h_{j\to a}\bigr)
\)
for all $j\in\partial a$ and form all leave-one-out products $\prod_{j\in\partial a\setminus\{i\}} t_{j\to a}$ efficiently using prefix/suffix products in $\mathcal{O}(|\partial a|)$ time per clause (rather than $\mathcal{O}(|\partial a|^2)$ by recomputing a product for each $i$). Finally, for numerical stability we clip all LLR-like quantities (fields and messages) to a fixed range and lower-bound the argument of the logarithm in \eqref{eq:paper_clause_to_var_iter} away from zero; these safeguards do not alter the intended fixed-point equations but prevent overflow/underflow in finite precision.
\subsection{Continuous variant (used inside continuous diffusion)}
\label{app:bp_continuous}

The continuous BP denoiser acts on the same $k$-SAT factor graph and uses the same
clause-to-variable BP updates as in the discrete case. At diffusion step $\ell$, the current
observation is $\bY_\ell\in\R^N$, and the goal is to approximate the denoiser
$\bm(\bY_\ell;\ti_\ell)$ associated with the tilted measure
$\mu_G^{\bY_\ell,\ti_\ell}$ from Eq.~\eqref{eq:Tilted}.

For diffusion level $\ti_\ell\in(0,1)$, we form the variable field

\begin{equation}
H_i(\ell)
\;=\;
\frac{\sqrt{\ti_\ell}}{1-\ti_\ell}\,Y_{\ell,i}
\;+\;
\sum_{b\in\partial i} s_{b,i}\,u_{b\to i}(\ell).
\label{eq:cont_field}
\end{equation}
and compute cavity fields on edges via the same identity as in the discrete case,
\begin{equation}
h_{i\to a}(\ell)
\;=\;
H_i(\ell) - s_{a,i}\,u_{a\to i}(\ell).
\label{eq:cont_cavity_identity}
\end{equation}
Given $\{h_{j\to a}(\ell)\}_{j\in\partial a\setminus i}$, clause-to-variable messages are then updated
using Eq.~\eqref{eq:paper_clause_to_var_iter} (with the same leave-one-out products, clipping, and other numerical safeguards as described in
Section~\ref{appendix:BP_discrete}). Iterating this update for $r$ rounds yields final fields $H_i(\ell)$ and magnetizations
$m_i(\ell)=\tanh(H_i(\ell))$, which serve as the BP-based continuous denoiser output at step $\ell$.

\subsection{Belief propagation in $k$-XORSAT }

\label{app:bp_intro}

Similarly to the previous section, in this section we describe the belief propagation (BP) / message passing algorithm that we use as a \emph{denoiser} inside both our \emph{discrete} and \emph{continuous} diffusion samplers to generate assignments satisfying random $k$-XORSAT formulas.
\paragraph{Problem definition and notation.}
A $k$-XORSAT instance consists of $M$ parity constraints over $N$ binary variables. We use the spin encoding
\(
\bx=(x_1,\dots,x_N)\in\{+1,-1\}^N
\),
and write the instance as a factor graph with variable nodes $i\in[N]$ and clause (factor) nodes $a\in[M]$. Each clause $a$ involves exactly $k$ variables with indices
\(
\partial a=\{i_a(1),\dots,i_a(k)\}\subseteq[N]
\),
sampled uniformly at random without replacement, and has a random sign (parity)
\(
s_a\in\{+1,-1\}.
\)
The constraint enforced by clause $a$ is
\begin{equation}
\prod_{i\in\partial a} x_i \;=\; s_a\, .
\end{equation}
Equivalently, in the Boolean encoding $\{0,1\}$ with $x_i=(-1)^{\tilde x_i}$, each constraint is a linear equation modulo $2$,
\(
\tilde x_{i_a(1)}\oplus\cdots\oplus \tilde x_{i_a(k)}=b_a
\),
where $s_a=+1$ if $b_a=0$ and $s_a=-1$ if $b_a=1$.

For a variable $i$, we denote by
\(
\partial i=\{a\in[M]: i\in\partial a\}
\)
the set of clauses incident to $i$. Throughout, messages are passed along directed edges of the factor graph: we write $u_{a\to i}$ for clause-to-variable messages and $h_{i\to a}$ for variable-to-clause messages. The corresponding \emph{local field} (or belief field) at variable $i$ is denoted by $H_i$, and its associated magnetization by
\(
m_i=\tanh(H_i)
\)
(in the continuous formulation).

\paragraph{Discrete BP message passing.}
In the discrete (hard) variant, messages take values in the three-point set
\(
\{+\infty,0,-\infty\}.
\)
We denote by $h_{i\to a}^{(r)}$ the variable-to-clause message and by $u_{a\to i}^{(r)}$ the clause-to-variable message at BP iteration $r$. Given a partial reveal vector $\by=(y_1,\dots,y_N)$ with
\(
y_i\in\{+\infty,0,-\infty\}
\)
(where $y_i=\pm\infty$ if variable $i$ is revealed and $y_i=0$ otherwise), the BP updates are
\begin{align*}
u_{a\to i}^{(r+1)}
&= s_a \prod_{j\in \partial i\setminus a} h_{j\to a}^{(r)}\, ,
\\
h_{i\to a}^{(r)}
&= \sign\!\Big( y_i + \sum_{b\in \partial i\setminus a} u_{b\to i}^{(r)} \Big)\, .
\end{align*}
The ``$\sign$ of an $\infty$-valued sum'' is understood with the convention that the sum never contains both $+\infty$ and $-\infty$ simultaneously (hence it is well-defined), and returns $0$ when all contributing terms are $0$.
After $r_{\BP}$ iterations, the final local field is computed as
\begin{equation}
H_i \;=\; \sign\!\Big( y_i + \sum_{b\in \partial i} u_{b\to i}^{(r)_{\BP}} \Big)\, ,
\end{equation}

\paragraph{Continuous BP message passing.}
In the continuous (soft) variant, messages are real-valued and we again denote by
$h_{i\to a}^{(r)}\in\mathbb{R}$ the variable-to-clause messages and by $u_{a\to i}^{(r)}\in\mathbb{R}$
the clause-to-variable messages at BP iteration $r$.
Given the current diffusion state $\by\in\mathbb{R}^N$ at time $\ell$ (the ``noisy observation''),
the BP updates are
\begin{align}
u_{a\to i}^{(r+1)} &={\rm atanh}\Big(s_a\prod_{j\in \partial i\setminus a}\tanh(h_{j\to a}^{(r)})\Big)\\
h_{i\to a}^{(r)}& = \lambda_{\ell}\,y_i+\sum_{b\in \partial i\setminus a} u^{(r)}_{b\to i}
\end{align}
After $r$ iterations, we form the (approximate) local field
\begin{equation}
H_i \;=\; \lambda_{\ell}\, y_i + \sum_{b\in\partial i} u_{b\to i}^{r_{\BP}}\, ,
\end{equation}
and the corresponding magnetization estimate
\begin{equation}
m_i \;=\; \tanh(H_i)\in[-1,1]\, .
\end{equation}

\emph{Denoiser in the continuous diffusion.}
In our continuous diffusion sampler, for each reverse step we run the above BP recursion on the factor graph
with input $\by$ and time-dependent scale $\lambda_{\ell}$ (determined by the diffusion schedule),
and we use the resulting magnetizations $\bm=(m_i)_{i\le N}$ as a denoising estimate of the underlying spins
(i.e., as the BP-based ``denoiser output'' plugged into the reverse update).
Equivalently, $H_i$ plays the role of a BP-informed score/field from which the denoiser $\tanh(H_i)$ is obtained.

\paragraph{Implementation details.}
We implement both denoisers as BP on the $k$-XORSAT factor graph.

\smallskip
\noindent\emph{Discrete BP (messages in $\{+\infty,0,-\infty\}$).}
We store directed messages in Python dicts keyed by edges:
\texttt{h[(i,a)]} for $h_{i\to a}$ and \texttt{u[(a,i)]} for $u_{a\to i}$, using adjacency lists
\texttt{var\_to\_clauses} and \texttt{clause\_to\_vars}.
Clause-to-variable updates compute $u_{a\to i}$ as the signed product of incoming $\sign(h_{j\to a})$
over $j\in\partial a\setminus i$, outputting $0$ if any incoming message is $0$.
Variable-to-clause updates use an ``$\infty$-safe'' $\sign(\sum)$ routine: it returns $+\infty$
(resp.\ $-\infty$) if only $+\infty$ (resp.\ only $-\infty$) appears among incoming messages,
and $0$ otherwise.

\smallskip
\noindent\emph{Continuous BP (real-valued messages).}
For speed we flatten edges into arrays and store the factor graph in CSR-like form. Clause updates compute
$u_{a\to i}=\atanh\!\big(s_a\prod_{j\in\partial a\setminus i}\tanh(h_{j\to a})\big)$
using prefix/suffix products to avoid division-by-zero, clipping both fields and the $\atanh$
argument for numerical stability; optional early-stop tolerance control convergence.

\smallskip
\noindent\emph{Use in diffusion.}
In both diffusion samplers, BP provides the denoiser at each reverse step: discrete BP is run with
partially revealed variables ($y_i\in\{\pm\infty,0\}$), while continuous BP takes the current noisy
${\bf y} \in\mathbb{R}^N$ and uses $\lambda_{\ell}=\sqrt{\ti_{\ell}}/(1-\ti_{\ell})$ to produce magnetizations
$m_i=\tanh(H_i)$ that enter the reverse diffusion update.

\subsection{Belief propagation in NAE-SAT}
\label{appendix:BP_naesat}


A NAE-$k$-SAT clause uses the same literal notation as in
Eq.~\eqref{eq:ksat_clause_paper_notation}, namely
$z_{i_a(j)}=s_{a,j}x_{i_a(j)}$. The difference from $k$-SAT is only in the
clause constraint: clause $a$ is satisfied when its literals are not all equal.
Equivalently, the two unsatisfying assignments are the all $+1$ and all $-1$
literal configurations.

The factor graph, variable-to-clause messages,
initialization, marginal computation, and other implementation details  are the same as in
Appendix~\ref{appendix:BP_discrete}. The only change is the clause-to-variable
update. For a clause $a$ and a variable $i\in\partial a$, the message
$u^{(r+1)}_{a\to i}$ is updated as:

\begin{align}
u^{(r+1)}_{a\to i}
&=
\frac12 \log\!\left(
1 - \frac{(1-\epsilon)}{2^{k-1}}\,
\prod_{j\in\partial a\setminus i}
\bigl(1+\tanh\!\bigl(s_{a,j}h^{(r+1)}_{j\to a}\bigr)\bigr)
\right)
\nonumber\\
&\phantom{AAA}
-\frac12\log\!\left(
1 - \frac{(1-\epsilon)}{2^{k-1}}\,
\prod_{j\in\partial a\setminus i}
\bigl(1-\tanh\!\bigl(s_{a,j}h^{(r+1)}_{j\to a}\bigr)\bigr)
\right).
\label{eq:paper_nae_clause_to_var_iter}
\end{align}

\subsection{Belief propagation in hypergraph bicoloring}
\label{appendix:BP_hypgraph_bicoloring}


A $k$-hypergraph bicoloring instance is the unsigned special case of
NAE-$k$-SAT. Each variable $x_i\in\{\pm1\}$ denotes the color of vertex $i$, and
each hyperedge $a$ is satisfied when the variables in $\partial a$ are not all
equal. Equivalently, the two unsatisfying assignments are the all $+1$ and
all $-1$ color configurations on the hyperedge. This is exactly NAE-$k$-SAT with $s_{a,i}=+1$ for every edge $(a,i)$. The factor graph,
variable-to-clause messages, initialization, marginal computation, and
implementation details are the same as in Appendix~\ref{appendix:BP_naesat}.

%
%
\section{Computing phase transition thresholds 
$\alpha_{\diff}$ and $\alpha_{\mask}$} \label{appendix:AlphaDiffMaskDerivationXORSAT}

In this section we describe the theoretical predictions
for $\alpha_{\diff}$ and $\alpha_{\mask}$, following \cite{montanari2007solving}.
In short, these phase transitions are identified with a `reconstruction'
(or `extremality') phase transition for a Gibbs measure on an infinite tree that is identified as the local weak limit of the uniform measure $\mu_G$.
We emphasize that the definitions given here are rigorous, 
but establishing that they actually correspond to a phase transition in the generative algorithm remain an open problem.
Rigorous results can be established for XORSAT 
under discrete diffusions (the result in that case is an easy 
modification of standard results \cite{dubois20023,ibrahimi2015set}),
and the spin glass models of \cite{huang2024sampling}.

\subsection{Tree models and local weak convergence}
\label{sec:TreeModel}

Under any of the random CSPs ensembles studied in this paper 
($k$-SAT, $k$-XORSAT, $k$-NAESAT, hypergraph bicoloring) the random
factor graph $G=(V,F,E)$ converges (in the sense of local weak convergence,
as $N,M\to\infty$ 
with $N/M\to\alpha$) to a rooted infinite tree formula
$T=(V_T,F_T,E_T)$ with root $o$, which is described next.
$T$ is a bipartite infinite tree with two type of vertices
`variable nodes' $V_T$ and `factor nodes' $F_T$, connected by edges 
$E_T$. The tree is rooted at a variable node $o$.
Each variable node is connected at the next level of the tree
to Poisson$(k\alpha)$ factor nodes (hence has total degree
Poisson$(k\alpha)+1$, with the exception of the root, which has degree 
Poisson$(k\alpha)$). 
Each factor node is connected to $k-1$
variable nodes at the next level (hence has total degree $k$)
We denote by $T(L)$ the finite subtree obtained by truncating the above after $L$ generations of variable nodes (in particular, $T(0)$ 
is the trivial tree that only contains the root).

To each variable node $i\in V_T$, we associate a variable $x_i\in\{+1,-1\}$, and write $\bx=(x_i:i\in V_T)$. To each factor node $a\in F_T$ we associate
a constraint $\psi_a(\bx_{\da})$ which we draw uniformly at random 
among all those of the CSP under study. With a slight
abuse of notation, we denote by $G$ the factor graph, together with the constraints $(\psi_a:a\in F_T)$. We can therefore speak of
a configuration $\bx$ satisfying (or not) the 
(infinite) formula $G$.

The free boundary Gibbs measure $\mu^{\sfree}_T$ is a way of defining 
a probability measure over solutions of the infinite formula $G$, which plays an important role for us.
Letting  $\mu_{T(L)}$ denote the uniform measure 
over solutions of the sub-formula $T(L)$ (extended arbitrarily beyond
level $L$), we obtain $\mu^{\sfree}_T$ by taking the weak limit of
$\mu_{T(L)}$ as $L\to\infty$. 
Existence of the limit (along a subsequence) 
is guaranteed by standard compactness arguments.

We then imagine sampling from $\bx_*\sim\mu^{\sfree}_T$, and revealing information
$\by=(y_i: i\in V_T)$ where in continuous diffusions we have 
$\by=\sqrt{\ti}\bx_*+\sqrt{1-\ti}\, \bg$, for  
$\bg = (g_i)_{i\in V_T} \sim \normal(0,\id_{V_T})$ and in
 discrete masked diffusions we have
 $y_i=x_{*,i}$ independently  with probability $\tdi$ for each $i\in V_T$, 
 and $y_i=*$ otherwise.
 
 We then define  $\mu_{T}^{\by,\ti}(\bx)$ to be the 
 conditional distribution of $\bx$ given $\by$ under the above joint distribution.
 Equivalently, this can be defined by 
 by taking the limit $L\to\infty$ of the measures
 \begin{align}
    \mu_{T(L)}^{\by,\ti}(\bx)& :=
     \frac{1}{Z_{T(L)}}\prod_{i\in V_{T(L)}} 
     e^{\sqrt{\ti}y_ix_i/(1-\ti)}
     \prod_{a\in F_{T(L)}}\psi_a(\bx_{\partial a})\, ,\\
     \mu_{T(L)}^{\by,t}(\bx) &:=
     \frac{1}{Z_{T(L)}}\prod_{i\in V_{T(L)}} 
     \bfone_{y_i=x_{i}}
     \prod_{a\in F_{T(L)}}\psi_a(\bx_{\partial a})\, ,
 \end{align}
 respectively for continuous and discrete diffusions.

\subsection{Reconstruction/extremality phase transition and 
$\alpha_{\diff}$, $\alpha_{\mask}$}
\label{sec:GeneralMethodThreshold}

Let $\bx_{\ge \ell}:=(x_i:d_T(o,i)\ge \ell)$ denote the variables at level $\ell$ or below in $T$. 
In words, the reconstruction/extremality question 
for the Gibbs measure  $\mu^{\sfree}_T$ asks whether observing variables
$\bx_{\ge \ell}$ 
increases the chance of guessing correctly $x_{o}$
by an amount that is bounded away from zero as $\ell\to\infty$.
If this happens, we say that reconstruction is possible (and the measure is non-extremal): this is a measure of how correlated is the tree Gibbs measure. Since \cite{mezard2006reconstruction,krzakala2007gibbs}, it was realized that
this phase transition coincides with the statistical physics prediction for the dynamical phase transition $\alpha_{\sd}(k)$. 
Again, while this identification is fully rigorous, the fact that 
the physics prediction for $\alpha_{\sd}(k)$ actually coincided with 
shattering in the space of solutions and hardness of sampling has only 
been shown in certain cases, see e.g.
\cite{ibrahimi2015set,ayre2019hypergraph,nam2022one,nam2024one}.

The reconstruction phase transition can be analogously defined for
the conditional measure $\mu^{\by,t}_T$.
 Formally, we define the reconstruction parameter by
 \begin{align}
    R(t,\alpha):=
    \lim_{\ell\to\infty}
    \E\Big\|\mu_{T}^{\by,t}\big(x_{o}\in\,\cdot\,\big|\bx_{\ge \ell}\big)
    -\mu_{T}^{\by,t}\big(x_{o}\in\,\cdot\,\big)\Big\|_{\sTV}\, .
    \label{eq:RTdef}
 \end{align}
As mentioned above, $R(t,\alpha)$ measure how much better we can reconstruct the root value
$x_o$ using the information $\bx_{\ge \ell},\by$ than just using 
 $\by$.

We then define (this definition applies both to 
discrete and continuous diffusions, depending on 
which of the two measures $\mu^{\by,t}_T$ is used):
\begin{align}
\alpha_{\mask/\diff}(k):=\sup\big\{\alpha: \, R(t,\alpha)=0\forall 
t\in[0,1]\big\}\, .\label{eq:AlphaThreshold}
\end{align}
 The limit quantity $R(t,\alpha)$
 (and hence the thresholds) can be characterized by a distributional
 recursion which we can either study analytically
 or approximate numerically.

 In the next sections we provide further details about these computations.

\subsection{Computation of the threshold $\alpha_{\mask}$ for XORSAT}

For XORSAT and discrete diffusion, the
calculation of $R(t,\alpha)$ reduces to 
studying a simple one dimensional recursion for the quantity $q_{\ell}$,
which is the probability that the root value $x_o$
 is  completely determined  given the information $\bx_{\ge \ell},\by$.
The recursion for $q_{\ell}$ is given by (see e.g. \cite{MezardMontanari})
\begin{equation}
q_{\ell} \;=\; t + (1-t)\Big(1-e^{-k\alpha q_{\ell}^{k-1}}\Big)
\;=\; 1-(1-t)e^{-k\alpha q_{\ell}^{k-1}} \, .
\end{equation}
Denote by $q^z_{\ell}$ the sequence thus defined with initialization
$q^z_{0}=z$. The resulting threshold is
\begin{equation}
\alpha_{\mask} \;=\; \sup\big\{\alpha\in\R_{\ge 0}:\;
 \lim_{\ell\to\infty} (q^1_{\ell}-q^0_{\ell})=0\, \forall t\in [0,1]\big\}\, .
\end{equation}
Equivalently, it is the largest $\alpha$ such that the 
iteration has a unique fixed point $q\in[0,1]$:
\begin{equation}
    q \;=\;  1-(1-t)e^{-k\alpha q^{k-1}} \, .
    \label{eq:mask_fp}
    \end{equation}
    At the critical point, the fixed point becomes \emph{degenerate}, 
    i.e.\ $f(x)=x$ has a triple contact: $f(x)=x$, $f'(x)=1$, and $f''(x)=0$.

Differentiating \eqref{eq:mask_fp} with respect to $q$ once and twice yields the two additional conditions
\begin{equation}
1 \;=\; (1-t)\,k\alpha\,(k-1)\,q^{k-2}\,e^{-k\alpha x^{k-1}} \, ,
\label{eq:mask_fp_prime}
\end{equation}
and
\begin{equation}
0 \;=\; (1-t)\,k\alpha\,e^{-k\alpha q^{k-1}}
\Big[(k-1)(k-2)\,q^{k-3} \;-\; (k-1)^2\,k\alpha\,q^{2k-4}\Big] \, .
\label{eq:mask_fp_second}
\end{equation}

Solving the system \eqref{eq:mask_fp}, \eqref{eq:mask_fp_prime}, 
\eqref{eq:mask_fp_second} we get
\begin{align}
x_* &\;=\; \frac{k-2}{k-1},
\qquad
1-t_* \;=\; \frac{e^{\frac{k-2}{k-1}}}{k-1}\, ,\\
\alpha_{\mathrm{mask}}(k)
&\;=\;
\frac{1}{k\,x_*^{\,k-1}}\cdot \frac{k-2}{k-1}
\;=\;
\frac{1}{k}\left(\frac{k-1}{k-2}\right)^{k-2},
\qquad (k\ge 3)\, .
\label{eq:alpha_mask_closed_form}
\end{align}

\subsection{Distributional recursion}

We next describe the distributional recursion characterization of the reconstruction parameter $R(t,\alpha)$ (and consequently $\alpha_{\diff}/\alpha_{\mask}$, per definition 
\eqref{eq:AlphaThreshold}). For the sake of concreteness, we
will focus on the case of $k$-XORSAT and continuous diffusions,
but the suitable generalizations can be provided. 

In general, the approach is to write a distributional recursion for the 
log likelihood appearing (implicitly) in Eq.~\eqref{eq:RTdef},
namely
\begin{align}
h_{\ell}^+:=\frac{1}{2}
\log\frac{\mu_{T}^{\by,t}\big(x_{o}=+1\big|\bx_{\ge \ell}\big)}
{\mu_{T}^{\by,t}\big(x_{o}=-1\big|\bx_{\ge \ell}\big)}\, ,
\;\;\;\;\;\;
h_{\ell}^0:=\frac{1}{2}
\log\frac{\mu_{T}^{t,\by_{\le\ell}}\big(x_{o}=+1\big)}
{\mu_{T}^{t,\by_{\le\ell}}\big(x_{o}=-1\big)}\, .
\end{align}
An important technical simplification for XORSAT 
is that ---by a symmetry argument--- we can assume that  
$s_a=+1$ (recall that $s_1$ is the sign defining the $a$-th constraints)
and that $\bx_{\ge \ell}=+\bfone$.

We denote by $Q^{+}_{\ell}$, $Q^0_{\ell}$ the laws of $h^{+}_{\ell}$,
$h^{0}_{\ell}$.
These satisfy the following distributional recursion (here we slightly abuse notation because --a priori--
 $Q_{\ell}$, $\hQ_{\ell}$ are only probability measures, not functions)
\begin{align}
\hQ_{\ell+1}(u)  &= 
\int \delta_{u-f(h_1^{k-1})} \prod_{i=1}^{k-1}\de Q_{\ell}(h_i)\, ,
\label{eq:RDE_1}\\
Q_{\ell}(h) &=\sum_{m\ge 0} \sP_{m}
\int\, \delta_{h-h_0-\sum_{i=1}^{m}u_i}
\prod_{i=1}^{m}\de \hQ_{\ell}(u_i)\, \de G_t(h_0)\, ,\label{eq:RDE_2}
\end{align}
where $\sP_m=\Poisson(k\alpha)$ and $G_t=\normal(t,t)$.
Further $f(h_1^{k-1})=\atanh(\tanh(x_1)\cdots\tanh(x_{k-1})$.

The distributions $Q^{+}_{\ell}$, $Q^0_{\ell}$ are
defined by the recursion \eqref{eq:RDE_1}, \eqref{eq:RDE_2}, with two different initializations:
$Q^0_0=\delta_0$, and $Q^+_0=\delta_{+\infty}$ (note that this makes sense
because $h\mapsto \tanh(h)$ can be extended by continuity to 
$[-\infty,+\infty]$).

Given the sequence of distributions $Q^{+}_{\ell}$ and $Q^0_{\ell}$,
the threshold $\alpha_{\diff}$ is characterized by
\begin{align}
    \alpha_{\diff} \;=\; \sup\Big\{\alpha\in\R_{\ge 0}:\;
    \lim_{\ell\to\infty} 
    \Big[\int \tanh(h) \de Q^{+}_{\ell}(h) - \int \tanh(h) \de Q^0_{\ell}(h) \Big]=0 \;\;\forall \ti\in [0,1]\Big\}\, .
\end{align}

\subsection{Numerical computation of the thresholds $\alpha_{\diff}/\alpha_{\mask}$}

We next describe a numerical algorithm that subsamples the about
iteration to estimate $\alpha_{\diff}$ \cite{mezard2001bethe,mezard2006reconstruction}.

\paragraph{Algorithm}
We keep two samples of size $N$ (say $N=10^4$), representing respectively $\hQ$ and $Q$:
\begin{align}
    \hcQ^0_N:=\{u_{1},\dots,u_N\},\;\;\;\;\;\;\;
    \cQ^0_N:= \{h_{1},\dots,h_N\}\, ,\\
     \hcQ^+_N:=\{u_{1},\dots,u_N\},\;\;\;\;\;\;\;
    \cQ^+_N:= \{h_{1},\dots,h_N\}\, .
\end{align}
The only difference is that $\hcQ^0_N,\cQ^0_N$  are initialized to be all $0$,
while $\hcQ^+_N,\cQ^+_N$  are initialized to be all $+A$, with $A$ a large positive number (e.g. $10$).
We will use  $\hcQ_N,\cQ_N$ to refer to either of the two initialization. The $0$ and $+$
samples are never mixed.

Then the following cycle is repeated for $T$ iterations:
\begin{enumerate}
\item Update $\hcQ_N$ as follows. 

For each $i\in\{1,\dots,N\}$:
\begin{enumerate}
    \item Sample indices $j(1),\dots, j(k-1)\sim_{iid}\Unif([N])$ (with replacement)
    \item For each $a\in\{1,\dots\,k-1\}$, set 
    \begin{itemize}
        \item $\hat{h}_a= h_{j(a)}$ (from the sample $\cQ_N$).
    \end{itemize}
    \item Replace $u_i\leftarrow f(\hat{h}_1^{k-1})$
\end{enumerate}
\item Update $\cQ_N$ as follows. 

For each $i\in\{1,\dots,N\}$:
\begin{enumerate}
    \item Sample $m\sim_{iid}\Poisson(k\alpha)$
    \item Sample indices $j(1),\dots, j(m)\sim_{iid}\Unif([N])$ (with replacement)
    \item Sample $h_{0,i}\sim \normal(t,t)$ (mean $t$, variance $t$).
    \item Replace 
    \begin{align*}
    h_i\leftarrow h_{0,i}+\sum_{a=1}^{m} u_{j(a)} 
    \end{align*}
    \end{enumerate}
\end{enumerate}

After these $T$ iterations, we are left with populations $\cQ^+_N$, $\cQ^0_N$.
The size can be increased by taking the union over iterations $T, T+1$, \dots $T+\Delta-1$
(thus obtaining populations of size $N\cdot\Delta$).
We compute the mean difference
\begin{align}
\Delta(t,\alpha):= \frac{1}{N}\sum_{i=1}^N\tanh h_i(\cQ^+)-\frac{1}{N}\sum_{i=1}^N\tanh h_i(\cQ^0)
\end{align}

For a given value of $\alpha$, evaluate $\Delta(\ti,\alpha)$ for $\ti\in[0,1]$ on a grid
$S\subseteq [0,1]$. 
Fixing a small tolerance $\Delta_0$, we estimate:
\begin{align}
\hat{\alpha}_{\diff}:=
\sup\big\{\alpha\in\R_{\ge 0}: \; \Delta(\ti,\alpha)\le \Delta_0 \;\;
\forall \ti\in S\big\}\, .
\end{align}

For $k$-XORSAT, the values of $\alpha_{\mask}$, $\alpha_{\diff}$, and $\alpha_{\sd}$ are reported in Table~\ref{tab:xorsat_thresholds}.

\begin{table}[t]
\centering
\caption{Threshold values for $k$-XORSAT.}
\label{tab:xorsat_thresholds}
\begin{tabular}{c|ccc}
\hline
$k$ & $\alpha_{\mathrm{mask}}$ &\;\;\; $\alpha_{\mathrm{diff}}$ \;\;\;& $\alpha_{\sd}$ \\
\hline
3  & 0.666667 & 0.736 & 0.818469 \\
4  & 0.562500 & 0.632 & 0.772280 \\
5  & 0.474074 & 0.531 & 0.701780 \\
6  & 0.406901 & 0.511 & 0.637081 \\
7  & 0.355474 & 0.396 & 0.581775 \\
8  & 0.315203 & 0.384 & 0.534997 \\
9  & 0.282944 & 0.366 & 0.495255 \\
10 & 0.256578 & 0.350 & 0.461197 \\
\hline
\end{tabular}
\end{table}

\begin{figure}[t]
    \centering
    \includegraphics[width=0.85\textwidth]{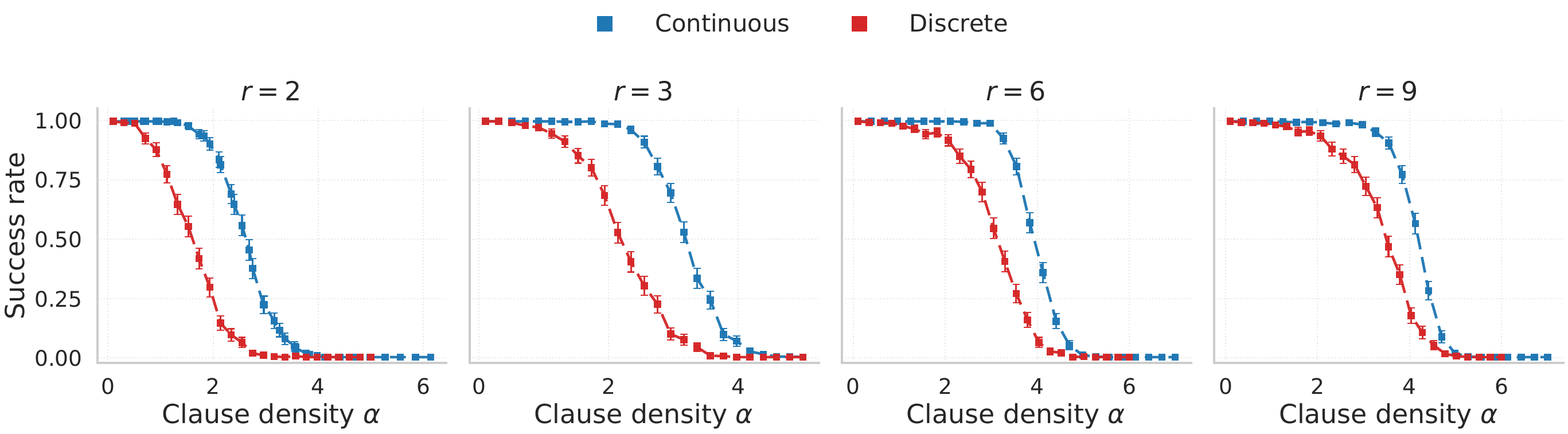}
    \caption{
        \textbf{4-NAE-SAT ($N=100$)}: Success rate as a function of clause density~$\alpha$ for continuous (blue) and discrete (red) diffusions using BP denoiser.
       Each panel corresponds to a different locality radius for the denoisers~$r \in \{2,3,6,9\}$, success rates are computed over 500 random formulas. The error bars are 95\% Wilson confidence intervals.
    }
    \label{fig:naesat_diffusion_vs_bp_n100}
\end{figure}

\begin{figure}[t]
    \centering
    \includegraphics[width=0.85\textwidth]{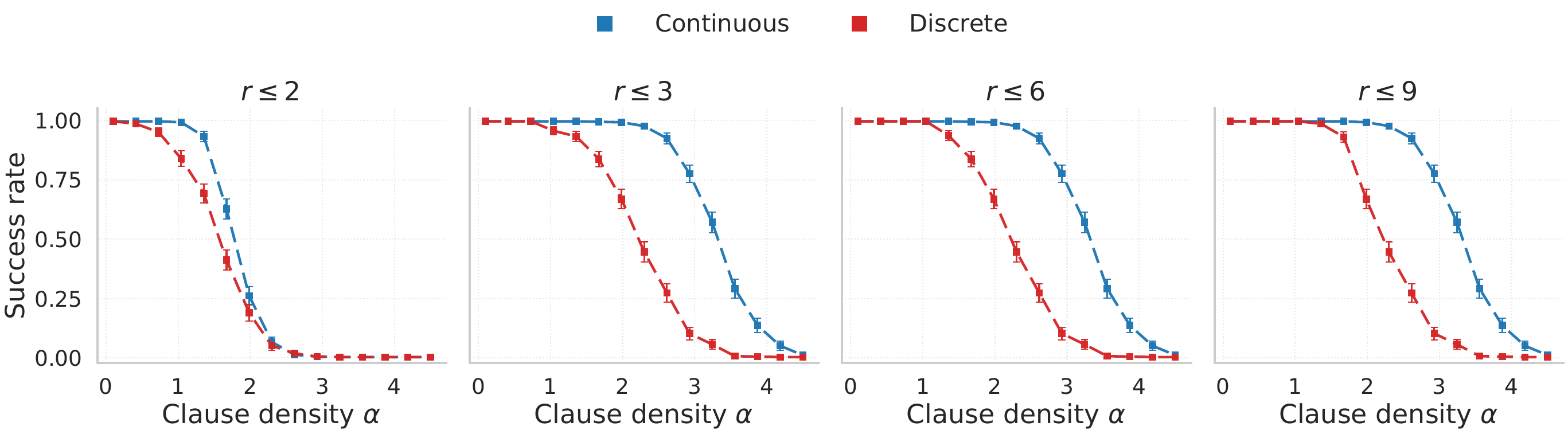}
    \caption{
        \textbf{4-Hypergraph-Bicoloring ($N=100$)}: Success rate as a function of clause density~$\alpha$ for continuous (blue) and discrete (red) diffusions using BP denoiser.
        Each panel corresponds to the optimal performance obtained with a locality radius less or equal than ~$r \in \{2,3,6,9\}$ (in this case we observed strong non-monotone behavior in $r$). Success rates are computed over 500 random formulas. The error bars are 95\% Wilson confidence intervals.
    }
    \label{fig:hyp_graph_diffusion_vs_bp_n100}
\end{figure}

\begin{figure}[t]
    \centering
    \includegraphics[width=0.97\linewidth]{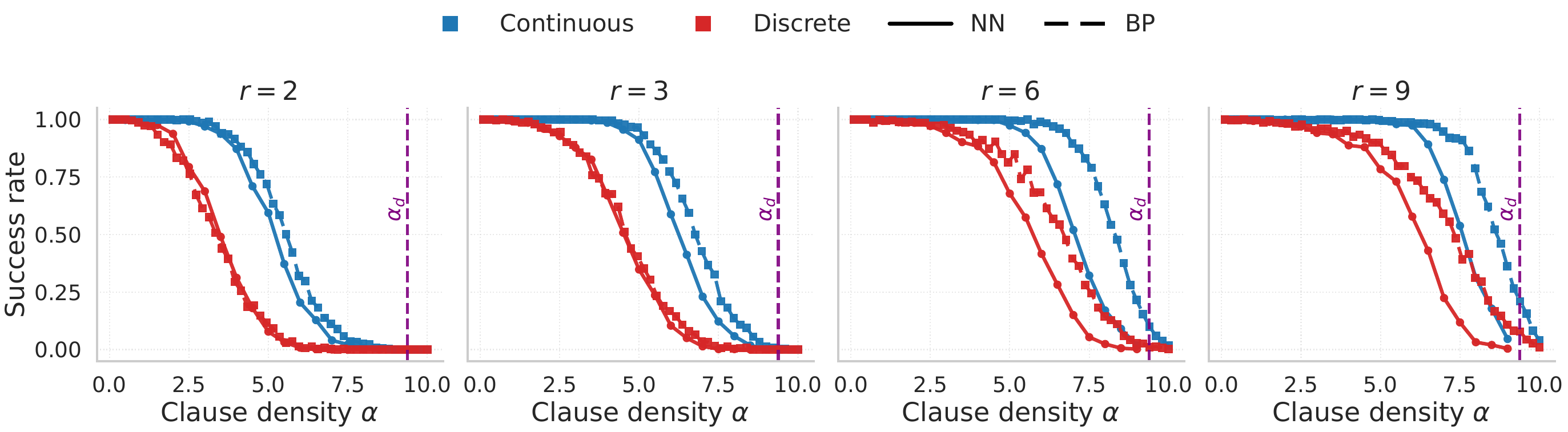}
    
    \caption{
        \textbf{4-SAT ($N=100$): learned NN denoiser (solid lines) vs.\ BP denoiser (dashed lines).}
        Success rate (probability of generating actual solutions) as a function of clause density~$\alpha$ for continuous (blue) and discrete (red) diffusions.
        Each panel corresponds to a different locality radius for the denoisers~$r \in \{2,3,6,9\}$.
        Success rates are computed over 500 random formulas.
        The vertical line marks the dynamical phase transition $\alpha_{\sd}(k=4) \approx 9.38$.
    }
    \label{fig:ksat_diffusion_vs_bp_n100}
\end{figure}

\begin{figure*}[t]
    \centering
    \includegraphics[width=\textwidth]{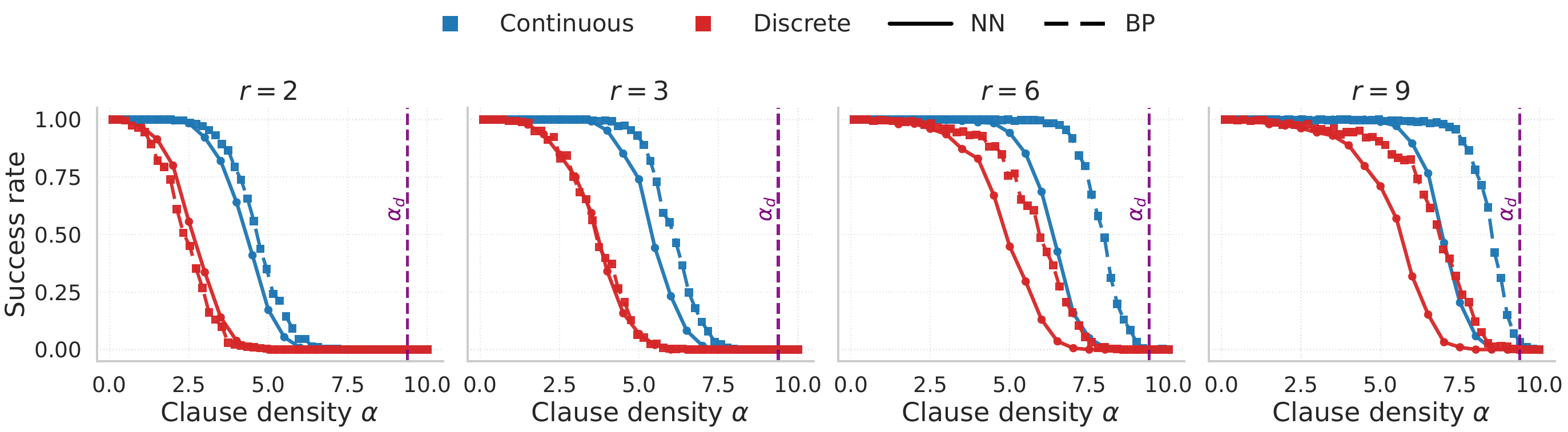}
    \caption{
        \textbf{4-SAT ($N=300$)}: learned NN denoiser (solid lines) vs.\ BP denoiser (dashed lines).
        Success rate as a function of clause density~$\alpha$ for continuous (blue) and discrete (red) diffusions.
        Each panel corresponds to a denoiser $r \in \{2,3,6,9\}$, success rates are computed over 500 random formulas.
        The vertical line marks the dynamic phase transition threshold $\alpha_{\sd} \approx 9.38$ for $k=4$.
    }
    \label{fig:ksat_diffusion_vs_bp_n300}
\end{figure*}

\begin{figure}[t]
    \centering
    \includegraphics[width=0.55\linewidth]{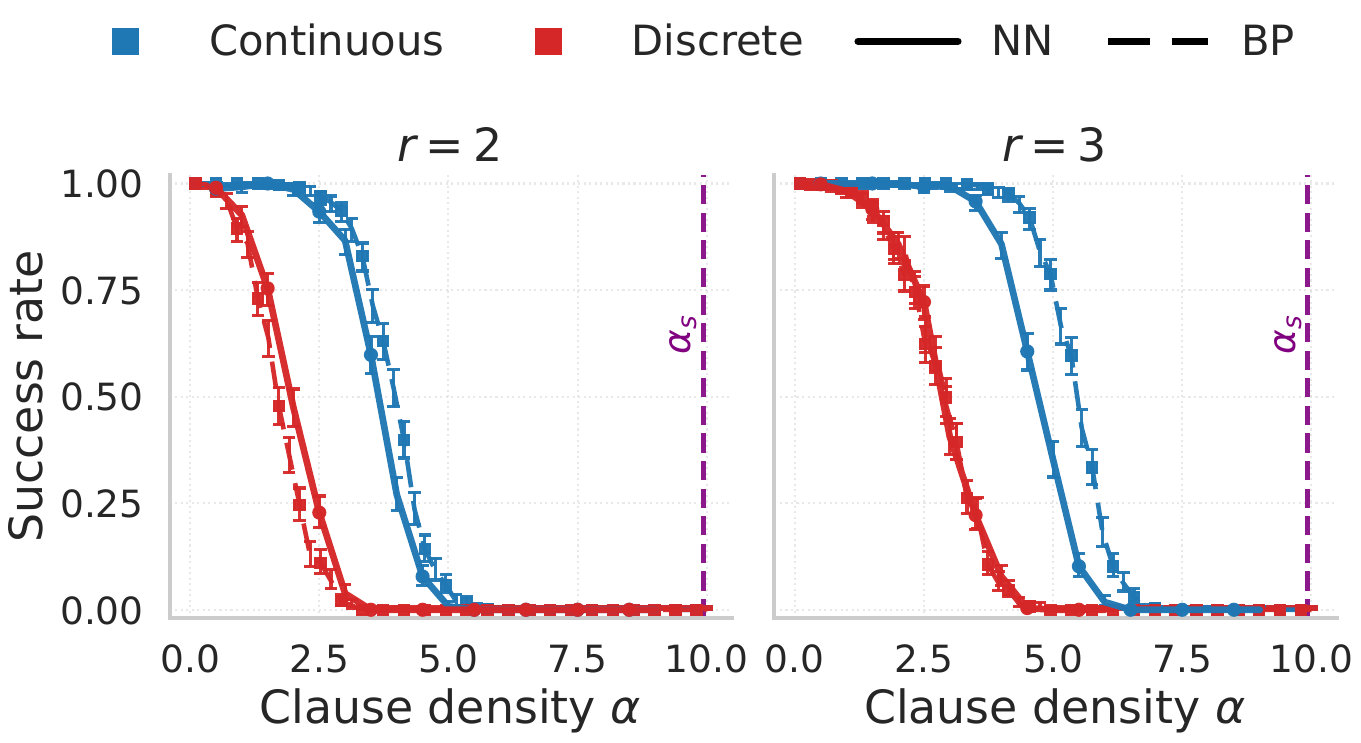}
    \caption{
        \textbf{4-SAT ($N=900$)}: learned NN denoiser (solid lines) vs.\ BP denoiser (dashed lines).
        Success rate as a function of clause density~$\alpha$ for continuous (blue) and discrete (red) diffusions.
        Each panel corresponds to a different locality radius for the denoisers $r \in \{2,3\}$, success rates are computed over 500 random formulas.
        The vertical line marks the dynamic phase transition threshold $\alpha_s \approx 9.38$ for $k=4$.
        The error bars are 95\% Wilson confidence intervals.
    }
    \label{fig:ksat_diffusion_vs_bp_n900}
\end{figure}

\begin{figure}[t]
    \centering
    \includegraphics[width=0.55\textwidth]{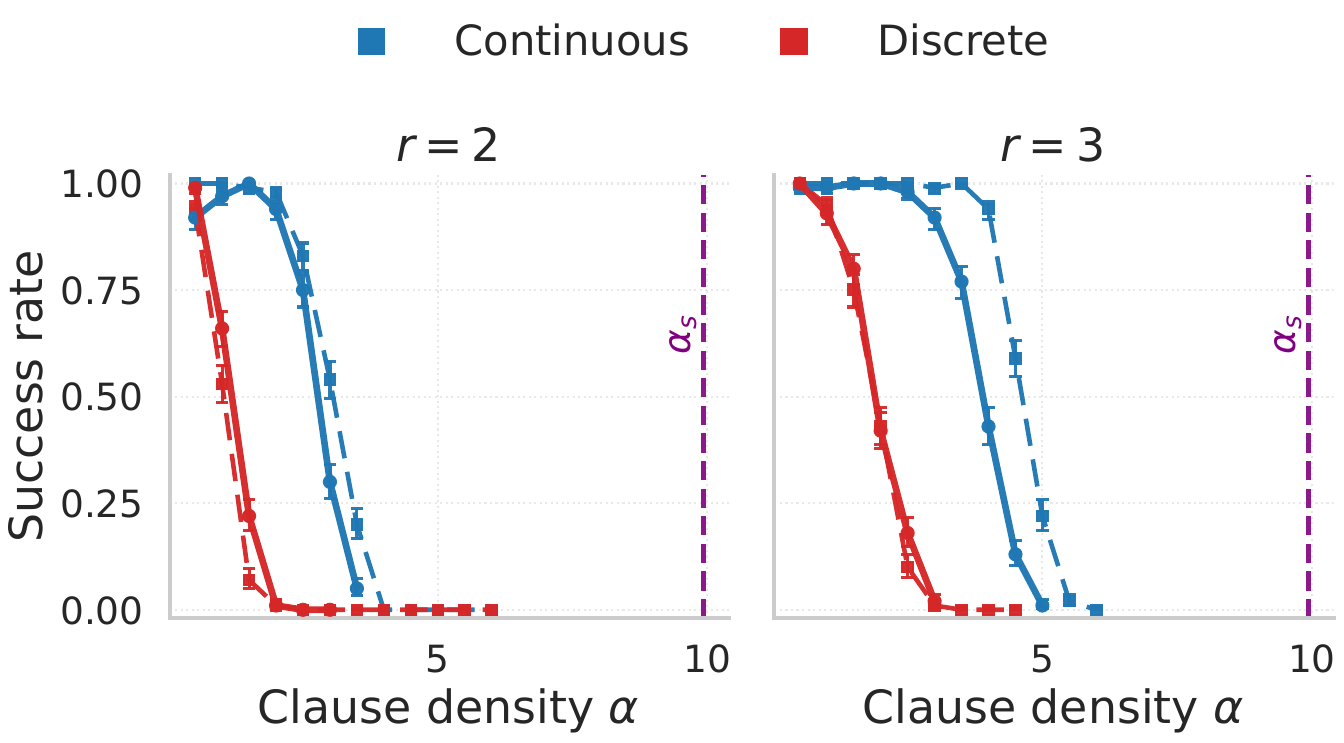}
    \caption{
        \textbf{4-SAT ($N=5000$)}: learned NN denoiser (solid lines) vs.\ BP denoiser (dashed lines).
        Success rate as a function of clause density~$\alpha$ for continuous (blue) and discrete (red) diffusions.
        Each panel corresponds to a different locality radius for the denoisers $r \in \{2,3\}$, success rates are computed over 500 random formulas.
        The vertical line marks the dynamic phase transition threshold $\alpha_s \approx 9.38$ for $k=4$.
        The error bars are 95\% Wilson confidence intervals.
    }
    \label{fig:ksat_diffusion_vs_bp_n5000}
\end{figure}


\begin{figure}[h]
    \centering
    \includegraphics[width=0.55\linewidth]{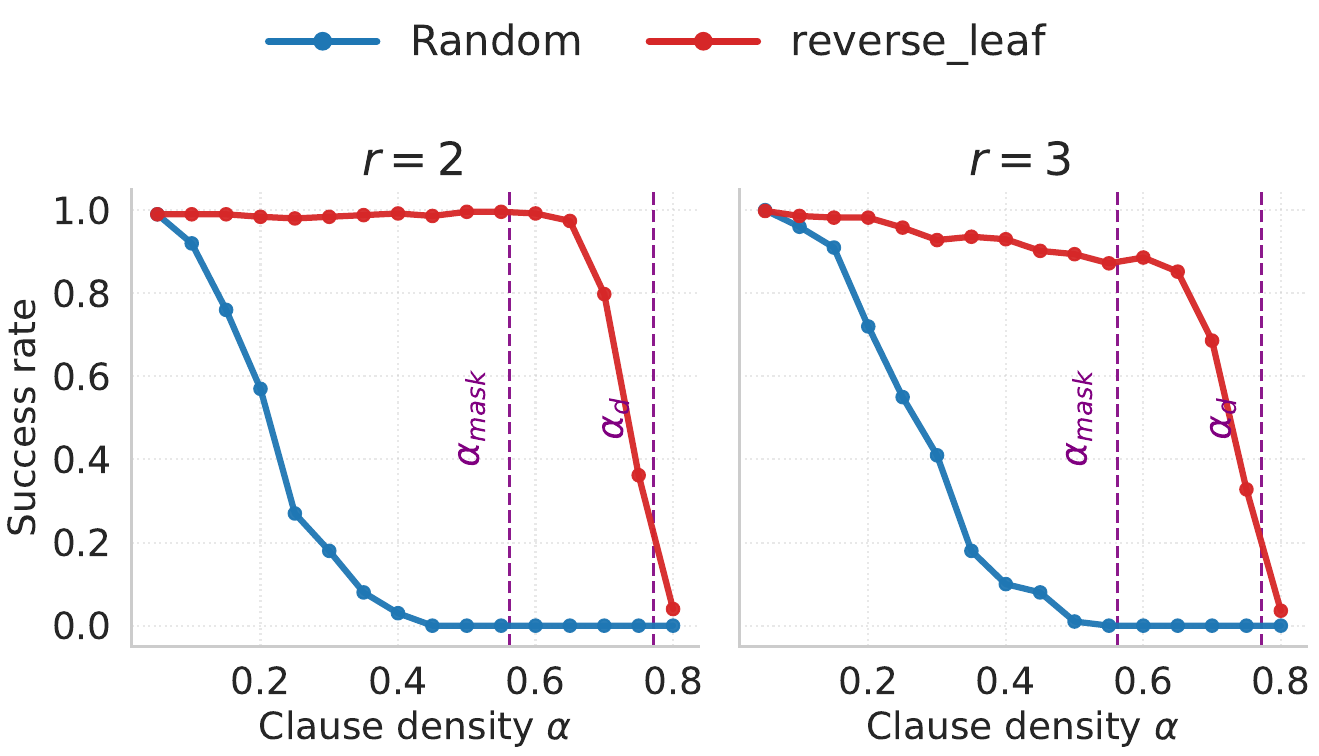}
    \caption{
       \textbf{4-XORSAT ($N=100$)}: reversed-leaf (red) vs random (blue) decoding ordering.
       Success rate as a function of clause density $\alpha$ obtained using \textbf{discrete} diffusion with learned NN denoiser. Each panel corresponds to a denoiser $r \in \{2,3\}$, success rates are computed over 500 random formulas. The two vertical lines are the theoretical thresholds for random ($\alpha_{\mask}$) and optimal ($\alpha_{\sd}$) decoding ordering.
    }
    \label{fig:xorsat_disc_nn_reverse_leaf}
\end{figure}

\begin{figure*}[h]
    \centering
    \includegraphics[width=\textwidth]{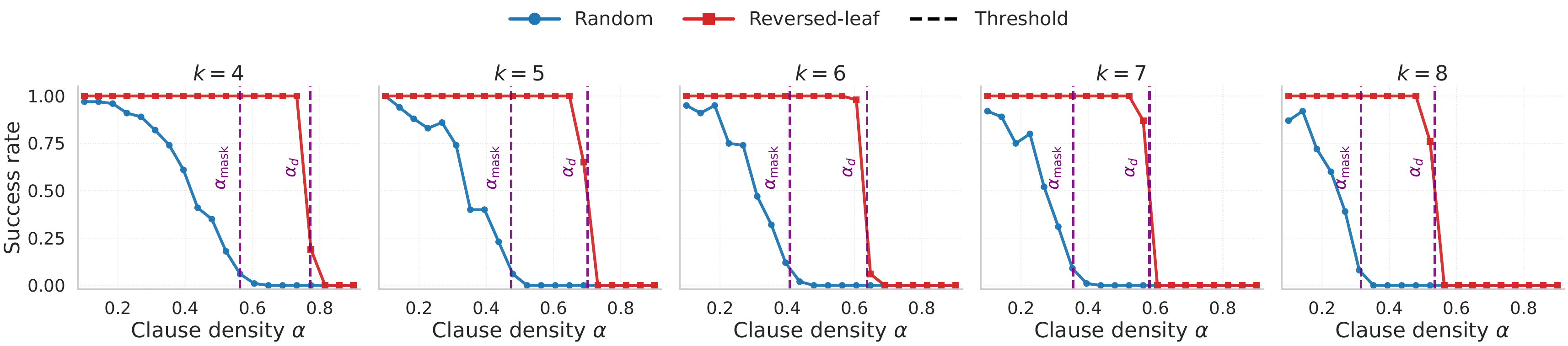}
    \caption{
        \textbf{$k$-XORSAT ($n=1000$)} : reverse-leaf (red) vs random (blue) decoding ordering.
        Success rate as a function of clause density $\alpha$ for $k\in\{4,\dots,8\}$ obtained using discrete diffusion with BP denoiser. The two vertical lines are the theoretical thresholds for random ($\alpha_{\mask}$) and optimal ($\alpha_{\sd}$) decoding ordering at every value of $k$.
    }
    \label{fig:disc_bp_n1000}
\end{figure*}


\section{Cavity initialization}
\label{appendix:CavityDetails}

We next describe the $\BPc$ initialization. 
To be definite, we focus on $k$-SAT but the definition is more general.
Recall from Appendix \ref{appendix:BPDetails} that  BP operates on messages that are associated to directed edges in $G$, 
each message being a probability distribution over $\{+1,-1\}$
or ---equivalently--- log-likelihood ratios. In  $\BPc(r)$ we initialize messages from variable notes $i\in V$ to factor nodes $a\in F$
$(\nu_{i\to a}(+1),\nu_{i\to a}(-1))$ 
(or equivalently, the log-likelihoods
$h_{i\to a}=(1/2)log\{\nu_{i\to a}(+1)/\nu_{i\to a}(-1)\}$.
to be i.i.d. with common distribution $Q$.

To define $Q$, recall the definition of free-boundary Gibbs measure 
on the infinite tree $T$, $\mu^{\sfree}_T$, defined in Section 
\ref{sec:TreeModel}. Then $Q$  is the distribution of the log-likelihood at the root under $\mu^{\sfree}_T$, i.e. the distribution
of 
\begin{align}
h_o:=\frac{1}{2}
\log\frac{\mu_{T}^{t,\by}\big(x_{o}=+1\big)}
{\mu_{T}^{t,\by}\big(x_{o}=-1\big)}\, .
\end{align}
It is standard to derive the following 
distributional equation for the $Q$ (involving the auxiliary distribution
$\hat{Q}$)
\begin{align}
\widehat{Q}(u)  &= 
\int \delta_{u-f(h_1^{k-1})} \prod_{i=1}^{k-1}\de Q(h_i)\, ,\\
Q(h) &= \frac{t}{2}\delta_{+\infty}(h)+\frac{t}{2}\delta_{-\infty}(h)+
(1-t)Q^0(h)\, ,\\
Q^0(h) &=\sum_{m_+,m_-\ge 0} \sP_{m_+}\sP_{m_-}
\int\, \delta_{h-\sum_{i=1}^{m_+}u_i^+ +\sum_{i=1}^{m_-}u_i^-}
\prod_{i=1}^{m_+}\de \widehat{Q}(u)(u_i^+)\,
\prod_{i=1}^{m_-}\de \widehat{Q}(u)(u_i^-)\, ,
\end{align}

In order to approximate the distribution $Q$, we use again a 
finite sample approximation described in Algorithm \ref{alg:PopDynCavityInitDiscrete}.
This yields two samples of size $N$, representing respectively $\widehat{Q}(u)$ and $Q^0$ :
\begin{align}
    \widehat{\mathcal{Q}}_N:=\{u_{1},\dots,u_N\},\;\;\;\;\;\;\;
    \mathcal{Q}_N:= \{h_1,\dots,h_N\}\, .
\end{align}

These empirical populations are used to initialize the $r=1$ BP denoiser (for discrete diffusions) as follows.
To estimate the marginal of a variable $x_i$, we run BP on a local neighborhood of $x_i$ for a single iteration.
For each directed variable-to-clause message $h_{j\to a}$, we set the initial condition according to whether $x_j$
has already been decoded:
\begin{itemize}
\item If $x_j$ is fixed, we initialize $h_{j\to a}=+\infty$ when $x_j=+1$ and $h_{j\to a}=-\infty$ when $x_j=-1$.
\item Otherwise, we initialize $h_{j\to a}$ by drawing an independent sample from the empirical population $\mathcal Q_N$.
\end{itemize}
We then compute the resulting belief at $i$ and sample $x_i$ from this marginal.

\begin{algorithm}[H]
\caption{Population Dynamics for Cavity Initialization (Discrete Diffusion)}
\label{alg:PopDynCavityInitDiscrete}
\begin{algorithmic}
\STATE {\bfseries Input:} clause size $k$; clause density $\alpha$; diffusion reveal rate $t\in[0,1]$; population size $N$; iterations $T$; (optional) pooling window $\Delta\ge 1$; update map $f(\cdot)$.
\STATE {\bfseries Output:} populations $\widehat{\cal Q}_N=\{u_1,\dots,u_N\}$ (approximating $\widehat Q(u)$) and $\mathcal{Q}_N=\{h_1,\dots,h_N\}$ (approximating $Q^0(h)$).

\STATE Initialize $h_i \gets 0$ and $u_i \gets 0$ for all $i\in\{1,\dots,N\}$.
\FOR{$\tau = 1,2,\dotsc,T$}
    \STATE {\bfseries Update $\widehat{\cal Q}_N$:}
    \FOR{$i = 1,2,\dotsc,N$}
        \STATE Sample indices $j(1),\dots,j(k-1)\sim_{\mathrm{i.i.d.}}\Unif([N])$ (with replacement).
        \FOR{$a=1,2,\dotsc,k-1$}
            \STATE Set $\hat h_a \gets +\infty$ w.p.\ $t/2$, $\hat h_a \gets -\infty$ w.p.\ $t/2$, and $\hat h_a \gets h_{j(a)}$ w.p.\ $1-t$.
        \ENDFOR
        \STATE $u_i \gets f(\hat h_{k-1})$.
    \ENDFOR

    \STATE {\bfseries Update $\cal Q_N$:}
    \FOR{$i = 1,2,\dotsc,N$}
        \STATE Sample $m_+,m_- \sim_{\mathrm{i.i.d.}} \mathsf{Poisson}(k\alpha/2)$.
        \STATE Sample indices $j_+(1),\dots,j_+(m_+) \sim_{\mathrm{i.i.d.}}\Unif([N])$ (with replacement).
        \STATE Sample indices $j_-(1),\dots,j_-(m_-) \sim_{\mathrm{i.i.d.}}\Unif([N])$ (with replacement).
        \STATE $h_i \gets \sum_{a=1}^{m_+} u_{j_+(a)} \;-\; \sum_{b=1}^{m_-} u_{j_-(b)}$.
    \ENDFOR
\ENDFOR

\IF{$\Delta>1$}
    \STATE (Optional) Pool the last $\Delta$ iterations to obtain $N\Delta$ samples for each population.
\ENDIF

\RETURN{$(\widehat{\cal Q}_N,{\cal Q}_N)$}
\end{algorithmic} \end{algorithm}

\section{Other CSPs}

Beyond $k$-SAT and $k$-XORSAT, we also evaluate on two additional classes of CSPs using the BP denoiser.

\textbf{NAE-SAT.}
Figure~\ref{fig:naesat_diffusion_vs_bp_n100} reports success rate results for $k$-Not-All-Equal SAT (NAE-SAT),
with $k=4$ (see Section \ref{appendix:BP_naesat} for definitions).
using both continuous and discrete diffusions and BP denoiser, at several values of the locality radius $r$.
As for SAT and XORSAT, we observe a significant advantage of continuous over discrete diffusiions. 

\textbf{Hypergraph bicoloring.}
 Figure~\ref{fig:hyp_graph_diffusion_vs_bp_n100} reports results for hypergraph bicoloring,
 with $k=4$ (i.e., every `hyperdge' comprises $4$ vertices), see Section \ref{appendix:BP_hypgraph_bicoloring} for definitions. 
 Again, we use  continuous and discrete diffusions and BP denoiser, at several values of the locality radius.
 One important observation is that, when using uniform initialization $\BPu(r)$ (which is what we do here) 
 we observe a non-monotonicity of accuracy in the number of BP iterations. We thus thake for each $r$ the optimal
 accuracy with number of iterations at most $r$.
Our main conclusion is confirmed here: continuous diffusions outperforms discrete diffusions.

\section{Further numerical results}
\label{app:further_results}

In this section, we report certain experimental results not presented in the main paper due to space limitations.

\textbf{$k$-SAT for larger $N$.}
In addition to the $k$-SAT results for $N=300$ with an $r=3$ local denoiser shown in Figure~\ref{fig:fig1_summary_ksat}, we report performance for several additional settings (:
\begin{itemize}
\item Figure~\ref{fig:ksat_diffusion_vs_bp_n300}: $N=300$ with $r\in\{2,3,6,9\}$;
\item  Figure~\ref{fig:ksat_diffusion_vs_bp_n100}: $N=100$ with $r\in\{2,3,6,9\}$;
\item Figure~\ref{fig:ksat_diffusion_vs_bp_n900}: $N=900$ with $r\in\{2,3\}$; 
\item nFigure~\ref{fig:ksat_diffusion_vs_bp_n5000}: $N=5000$ with $r\in\{2,3\}$. 
\end{itemize}
In each of these cases we report the success rate with continuous and discrete diffusions, both using BP and learned 
denoisers.

The results are remarkably consistent and confirm the conclusions of  Section~\ref{sec:results_cont_vs_disc}.
Continuous diffusion consistently outperforms discrete diffusion, and the learned local NN denoiser performs comparably to the BP denoiser. For $N=900$ (Figure~\ref{fig:ksat_diffusion_vs_bp_n900}) and $N=5000$ (Figure~\ref{fig:ksat_diffusion_vs_bp_n5000}), we additionally include $95\%$ Wilson confidence intervals, indicating that statistical uncertainty is negligible relative to the observed effect.

{\bf Reverse leaf ordering for $k$-XORSAT.} We carry out additional experiments with XORSAT, comparing
reversed leaf ordering and random ordering:
\begin{itemize}
\item  Figure~\ref{fig:disc_bp_n1000}: impact of ordering on discrete diffusion at $N=1000$ for
$k\in \{4,5,6,7,8\}$. (Figure~\ref{fig:disc_bp} in the main text focused on $N=300$, $k=4$.)
\item Figure~\ref{fig:xorsat_disc_nn_reverse_leaf}: impact of ordering on  discrete diffusion with an NN denoiser.
 (Figure~\ref{fig:disc_bp} in the main text used BP denoiser.)
 \end{itemize}
Again, we observe that the conclusions in the main text are confirmed by these additional experiments.

\section{Proofs}
\label{app:Analytical}

\subsection{Proof of Theorem \ref{thm:BPc}}

We first recall the definition of local (weak) convergence 
for Gibbs measures (see e.g. \cite{dembo2010gibbs}), adapting it to the
present setting. For concreteness, we consider $k$-SAT, but 
the case of general CSP can be treated analogously.

Recall the definition of `free boundary Gibbs measure' from 
Section \ref{sec:GeneralMethodThreshold}. 
In particular, let $T=(V_T,F_T,E_T)$ 
be the infinite tree SAT formula rooted at variable node $o\in V_T$, 
defined by the fact that the $L$-generation tree $T(L)$ of 
Section \ref{sec:Cavity} is obtained by taking the first $L$ 
generations of $T$.
HEre we regard the free boundary Gibbs measure as a joint law over 
$(T,\bx,\by)$ with $\bx$ a satisfying assignment of $T$.
The assumption of the theorem implicitly requires that the
the subsequential limit used to define this measure is unique.
We lift this to a joint law of $(T,\bx,\by)$, by letting 
$\by = \sqrt{\ti}\bx+\sqrt{1-\ti}\bg$, for $\bg = (g_i)_{i\in V_T}
\sim_{iid}\normal(0,1)$. We denote this joint law by $\P$
and its marginal on $T(r)$ by $\P_{T(r)}$

Let $I\sim \Unif([N])$, and let $\P_{N,r}$ be the joint law
of $(\Ball_G(I,r), \bx_{\Ball_G(I,r)},\by_{\Ball_G(I,r)})$ (were we view 
$\Ball_G(I,r)$ as rooted at $I$.
We say that the measure $\mu_G^{\by,\ti}$ converges locally to the 
free boundary Gibbs measure if, for every $r>0$, we have
$\P_{N,r}\Rightarrow \P_{T(r)}$, where $\Rightarrow$ denotes weak convergence.
Since $\hm_{\BPc(r)}(\,\cdot\, )$ is a bounded continuous function of
$\bY_{\Ball_G(I,r)} =(\Ball_G(I,r), \by_{\Ball_G(I,r)})$ (plus additional randomness for the initialization), we have
\begin{align}
 \lim_{N\to\infty}\E_N\big\{\big(x_I-\hm_{\BPc(r)}(\bY_{\Ball_G(I,r)})\big)^2\big\} 
=\E\big\{\big(x_o-&\hm_{\BPc(r)}(\bY_{T(r)})\big)^2\big\} \, .\label{eq:FirstStepThm1}
\end{align}
Further , the infimum on the right-hand side of Eq.~\eqref{eq:BPc-claim} can be restricted
to estimators $\hm$ such that $\sup_{\bY}|\hm(\bY)|\le 1$ 
(because $|x_I|\le 1$)
Define 
\begin{align}
\cM_{L,C}:=\Big\{\hm : \|\hm\|_{\infty}\le 1,\;  |\hm(\Ball, \by)-\hm(\Ball,\by')|\le L\|\by-\by'\|\;
\forall |\Ball|\le C;\;\forall \by,\by'\Big\}\, .
\end{align}
Using the fact that $\lim_{C\to\infty}\sup_N\P(|\Ball_G(i,r)| \ge C)=0$, and the fact that Lipschitz functions are dense in $L^2$, we conclude that for every $\eps$ there exists $L,C$ independent of $N$ such that
\begin{align}
\inf_{\hm\in\cM_{L,C}}  \E_N\big\{\big(x_i-\hm(\bY_{\Ball_G(I,r)})\big)^2\big\} \le   \inf_{\hm} \E_N\big\{\big(x_i-\hm(\bY_{\Ball_G(I,r)})\big)^2\big\} +\frac{\eps}{4}\,.
\end{align}
Let $\hm_N\in \cM_{L,C}$ an estimator that nearly achieves the infimum, i.e. such that 
\begin{align}
\E_N\big\{\big(x_i-\hm_N(\bY_{\Ball_G(I,r)})\big)^2\big\}\le 
\inf_{\hm\in\cM_{L,C}}  \E_N\big\{\big(x_i-\hm(\bY_{\Ball_G(I,r)})\big)^2\big\}+\frac{\eps}{4}\,.
\end{align}
Therefore using the last two displays we have:
\begin{align}
\E_N\big\{\big(x_i-\hm_N(\bY_{\Ball_G(I,r)})\big)^2\big\}\le 
\inf_{\hm}  \E_N\big\{\big(x_i-\hm(\bY_{\Ball_G(I,r)})\big)^2\big\}+\frac{\eps}{2}\,.
\end{align}
By Ascoli-Arzel\'a and a diagonal argument,  for any sequence $(N^0_j)$ there exists a further subsequence
$(N_j)\subseteq (N^0_j)$ and  $\hm_{\infty} \in\cM_{L,C}$ such that, for all $j\ge 1$
$\sup_{|\Ball|\le C,\|\by\|^2\le Cj }|\hm_{N_j}(\Ball, \by)-\hm_{\infty}(\Ball,\by)|\le \eps\cdot 2^{-j}$. Hence, using the previous equation,  for all $j$
\begin{align*}
\E_{N_j}\big\{\big(x_I-\hm_{\infty}(\bY_{\Ball_G(I,r)})\big)^2\big\}
\le &\inf_{\hm}\E_{N_j}\big\{\big(x_I-\hm_{\infty}(\bY_{\Ball_G(I,r)})\big)^2\big\}+
\frac{\eps}{2^{j}}\\
&+\frac{\eps}{2}
+4 \cdot \P\Big(\|\bY_{\Ball_G(I,r)}\|^2\ge Cj;|\Ball_G(I,r) |\le C\Big)\, .
\end{align*}
Hence, using standard tail bounds on chi-squared random variables, and eventually taking a further subsequence, we conclude that for any sequence $(N^0_j:j \ge 1)$, there exists a further subsequence
 $(N_j:j \ge 1)\subseteq (N^0_j:j \ge 1)$, such that
\begin{align}
\E_{N_j}\big\{\big(x_I-\hm_{\infty}(\bY_{\Ball_G(I,r)})\big)^2\big\}
\le \inf_{\hm}\E_{N_j}\big\{\big(x_I-\hm(\bY_{\Ball_G(I,r)})\big)^2\big\}+\frac{3}{4}\eps\, .
\end{align}
Taking the limit $j\to\infty$, and using the fact that the above holds for every initial
subsequence $(N^0_j)$, we get (notice that the limit of the left-hand side depends on the subsequence $(N_j)$ only through $\hm_{\infty}$):
\begin{align}
\E\big\{\big(x_o-\hm_{\infty}(\bY_{T(r)})\big)^2\big\}
\le \liminf_{N\to\infty}\inf_{\hm\in \cM_{L,C}}\E_{N}\big\{\big(x_I-\hm_{\infty}(\bY_{\Ball_G(I,r)})\big)^2\big\}+\frac{3}{4}\eps\, .
\end{align}
Therefore
\begin{align}
\inf_{\hm}\E\big\{\big(x_o-\hm(\bY_{T(r)})\big)^2\big\}
\le \liminf_{N\to\infty}\inf_{\hm\in \cM_{L,C}}\E_{N}\big\{\big(x_I-\hm_{\infty}(\bY_{\Ball_G(I,r)})\big)^2\big\}+\frac{3}{4}\eps\, .
\end{align}
The infimum on the left hand side is achieved by $\hm_{\BPc(r)}$ by construction and therefore by Eq.~\eqref{eq:FirstStepThm1}
\begin{align}
 \lim_{N\to\infty}\E_N\big\{\big(x_I-\hm_{\BPc(r)}(\bY_{\Ball_G(I,r)})\big)^2\big\} 
 \le \liminf_{N\to\infty}\inf_{\hm}\E_{N}\big\{\big(x_I-\hm_{\infty}(\bY_{\Ball_G(I,r)})\big)^2\big\}+\frac{3}{4}\eps\, .
\end{align}
This proves our claim.

\thispagestyle{empty}

\subsection{Proof of Theorem \ref{thm:LeafRemoval}}

It is convenient to replace the $\{+1,-1\}$ representation of variables by the $\{0,1\}$ representation,
with multiplications replaced by sum modulo two (we still use $\bx$ to denote the variables in this domain). 
A formula can therefore be represented by a matrix
$\bA \in \{0,1\}^{M\times N}$, where row $a\in\{1,\dots M\}$ has non-zero entries in correspondence to 
the indices  $i_1(a),\dots ,i_k(a)$ of variable participating constraint $a$, and a vector $\bb\in\{0,1\}^M$
which represents the signs $s_a$ (with $b_a=0$ if and only if $s_a=+1$).
In this representation, $\bx\in\{0,1\}^N$ is a solution if and only if  $\bA\bx = \bb\, \mod 2$.

We will assume that the leaf removal succeeds. Under this ordering the matrix $\bA$ is upper triangular,
with ones on the main diagonal:
\begin{align}
\bA = \Big[\bT\Big| \bB \Big]\, ,
\end{align}
where $\bT\in\{0,1\}^{M\times M}$ is square upper triangular, with ones on the diagonal, and $\bB\in\{0,1\}^{M\times (N-M)}$,
If we correspondingly decompose $\bx = [\bu|\bv]$ with $\bv\in \{0,1\}^{N-M}$, we have that each $\bv$ extends uniquely to 
a solution of the overall linear system.

Hence a uniformly random solution can be generated by letting $\bu\sim\Unif(\{0,1\}^{N-M})$,
and extending it uniquely. For $i\le M$, we let $x^*_i(\bu)$ be such unique extension. 
We next will describe this procedure in a different but obviously equivalent fashion.

We note that, in this notation, the reversed leaf ordering is (without loss of generality):
$\sigma(1) = N$, $\sigma(2)= N-1$,\dots $\ord(N) = 1$.
Hence a uniformly random solution can be generated
 by the following procedure:
 \begin{enumerate}
 \item[$(I)$] For $\ell\in\{0,1,\dots, N-M-1\}$, sample $\ox_{\ord(\ell+1)}\sim\hmu^{\ord,\ell}_G(\, \cdot\,)= \Unif(\{0,1\})$.
 \item[$(II)$] For $\ell\in\{N-M,1,\dots, N-1\}$, sample $\ox_{\ord(\ell+1)}\sim\hmu^{\ord,\ell}_F(\, \cdot\,)$,
 where $\hmu^{\ord,\ell}_F(x_{\ord(\ell+1)}=x^*_{\ord(\ell+1)}(\obx_{M+1:N}) )=1$. 
 \end{enumerate}
 Note that the last point is just an indirect way of saying that we set $\ox_{\ord(\ell+1)} =x^*_{\ord(\ell+1)}(\obx_{M+1:N})$.

 Comparing with the definition of masked discrete diffusions, we are only left with the task of showing that 
 the $\BPu(\infty)$ marginals coincide with the marginals $\hmu$ described above.
 Fix any $\ell\in\{0,\dots,N-1\}$.
 By the above construction, the system is always solvable 
 (under the assumption that leaf removal is successful), and therefore (by recasting $\bx$ as $\bx-\bx_0$ for $\bx_0$ a specific solution), 
 we can and will assume $\bb=0$. By the same argument, we can assume
 $\obx^{\ord}_{1:\ell} = \bzero_{\ell}$.

 Under these reductions the the analysis of $\BPu$ reduces to the
 analysis of BP decoding for a low-density parity check code under the binary
 erasure channel. We recall three elementary facts about this theory
 (see, e.g. \citep[Chapter 3]{RiU08} or \citep[Section 15.3]{MezardMontanari}):
 $(i)$~The BP estimate of the marginal of $x_i$ after any number $r$ of
 iterations is either 
 $\nu_i^{(r)} = (1,0)$  (where $\nu_i^{(r)} = (\nu^{(r)}_i(0),\nu^{(r)}_i(1))$) or
 $\nu^{(r)}_i=(1/2,1/2)$; $(ii)$~Estimates are `monotone' in the iteration number,
 namely if $\nu_i^{(r)} = (1,0)$, then  $\nu_i^{(r')} = (1,0)$ for all $r'\ge r$;
 $(iii)$~If $\nu_i^{(r)} = (1,0)$ at some $r$, then the actual conditional marginal of variable  $x_i$ is 
 $\mu_G^{\ord,\ell}(x_i=0|\obx^{\ord}_{1:\ell}=\bzero_{\ell})=1$.

 We are now in position to finish the argument:
 \begin{enumerate}
 \item[$(I)$] For $\ell\in\{-1,0,\dots, N-M-2\}$
 (hence $\ord(\ell+1)\in\{N,\dots,M+1\}$)
 we know that  $\mu^{\ord,\ell}(x_{\ord(\ell+1)}=0|\obx^{\ord}_{1:\ell}=\bzero_{\ell})=1/2$ 
 (because as explained above the marginal distribution on $\bx_{M+1,N}$
 is uniform on $\{0,1\}^{N-M}$) and therefore by property $(iii)$ above the 
 BP marginal is also $(1/2,1/2)$ for all iterations.
 \item[$(II)$] For $\ell\in\{N-M-1,\dots,N\}$
  (hence $\ord(\ell+1)\in\{M,\dots,1\}$) we know that (by the upper triangular structure of $\bA$, 
  variable $x_{\sigma(\ell+1)}$ appears in an equation of the form 
  $x_{\sigma(\ell+1)}+x_{j_2}+\dots +x_{j_k}=0,\, \mod 2$, where
  $j_2,\dots,j_k>\sigma(\ell+1)$, and hence we are conditioning to 
  $x_{j_2}=0$, \dots $x_{j_k}=0$. Hence after one iteration of BP (i.e. for $r\ge 1$),  $\nu^{(r)}_{\ord(\ell+1)}=(1,0)$.
 \end{enumerate}
 This proves that the $\BPu(\infty)$ estimate coincide with $\hmu$
 given above, hence completing the proof of the first part of the 
 statement.
Finally the `in particular' part of the statement follows immediately from the fact that the leaf removal procedure succeeds with high probability for $\alpha<\alpha_{\sd}(k)$.


\end{document}